\documentclass[runningheads]{llncs}

\usepackage{eccv}

\usepackage{eccvabbrv} 

\usepackage{graphicx}
\usepackage{amsmath}
\usepackage{amssymb} 
\usepackage{booktabs}

\usepackage{url}
\usepackage{amsfonts}       
\usepackage{nicefrac}       
\usepackage{microtype}      
\usepackage{algorithm}
\usepackage{algorithmic}
\usepackage{wrapfig}
\usepackage{epsfig}
\usepackage{multirow}
\usepackage{xcolor}

\usepackage{threeparttable}
\usepackage{colortbl}
\usepackage{enumitem}
\usepackage{siunitx}
\usepackage{bm}
\usepackage{bbm}
\usepackage[export]{adjustbox}
\usepackage{listings}
\usepackage{pifont}
\usepackage{dblfloatfix}
\usepackage{lipsum}
\newcommand{\cmark}{\ding{51}}%
\newcommand{\xmark}{\ding{55}}%

\newcommand{\ccnote}[1]{\textcolor{red}{#1}}

\definecolor{delay}{RGB}{230,94,42}
\definecolor{mywarning}{RGB}{233,144,61}
\definecolor{mygray}{gray}{.9}
\definecolor{mygray2}{gray}{.52}
\definecolor{ggray}{RGB}{127,127,127}
\definecolor{reda}{RGB}{192,0,0}
\definecolor{redb}{RGB}{217,148,143}
\definecolor{myyellow}{RGB}{190,144,0}
\definecolor{mygreen}{RGB}{80,100,40}
\definecolor{myblue}{RGB}{30,90,100}

\newcommand{\myhyperlink}[3][black]{\hyperlink{#2}{\color{#1}{#3}}}

\makeatletter
\newcommand{\thickhline}{
  \noalign {\ifnum 0=`}\fi \hrule height 1pt
  \futurelet \reserved@a \@xhline
}

\usepackage{arydshln} 

\usepackage{tabulary}
\newcolumntype{y}[1]{>{\raggedright\arraybackslash}p{#1pt}}
\newcolumntype{z}[1]{>{\raggedleft\arraybackslash}p{#1pt}}
\usepackage[pagebackref,breaklinks,colorlinks,citecolor=eccvblue]{hyperref}

\begin{document}

\title{\!\!\!\!\!\!\!\!General and Task-Oriented Video Segmentation\!\!\!\!\!\!\!\!}

\titlerunning{General and Task-Oriented Video Segmentation}

\author{Mu Chen\inst{1}\and
Liulei Li\inst{1}\and
Wenguan Wang\inst{2}\and
Ruijie Quan\inst{2}\and
Yi Yang\inst{2\raisebox{-2.5pt}{\thanks{Corresponding author: Yi Yang (yangyics@zju.edu.cn)}}}
}

\authorrunning{M. Chen et al.}

\institute{
    ReLER Lab, AAII, University of Technology Sydney, Australia \and
    ReLER Lab, CCAI, Zhejiang University, China \\
    \url{https://github.com/kagawa588/GvSeg}
}

\maketitle

\vspace{-15pt}
\begin{abstract}
{We present \textsc{GvSeg}, a \textbf{g}eneral \textbf{v}ideo \textbf{seg}mentation framework for addressing four different video segmentation tasks} (\ie, instance, semantic, panoptic, and exemplar-guided) while maintaining an identical architectural design. 
Currently, there is a trend towards developing general video segmentation solutions that can be applied across multiple tasks. This streamlines research endeavors and simplifies deployment. However, such a highly homogenized framework in current design, where each element maintains uniformity, could overlook the inherent diversity among different tasks and lead to suboptimal performance. To tackle this, \textsc{GvSeg}: \textbf{i)} provides a holistic disentanglement and modeling for segment targets, thoroughly examining them from the perspective of appearance, position, and shape, and on this basis, \textbf{ii)} reformulates the query initialization, matching and sampling strategies in alignment with the task-specific requirement.  
These architecture-agnostic innovations empower \textsc{GvSeg} to effectively address each unique task by accommodating the specific properties that characterize them. Extensive experiments on seven gold-standard benchmark datasets demonstrate that \textsc{GvSeg} surpasses all existing specialized/general solutions by a significant margin on four different video segmentation tasks.
\vspace{-10pt}
\end{abstract}

\section{Introduction}
\label{sec:intro}

Identifying target objects and then inferring their spatial locations over time in a pixel observation constitute fundamental challenges in computer vision\!~\cite{zhou2022survey}. Depending on discriminating unique instances or semantics associated with targets, exemplary tasks include: \textit{exemplar-guided} video segmentation (EVS) that tracks objects with given annotations at the first frame, video \textit{instance} segmentation (VIS), video \textit{semantic} segmentation (VSS), and video \textit{panoptic} segmentation (VPS) which entails the delineation of foreground instance tracklets, while simultaneously assigning semantic labels to each video pixel. 
Prevalent work primarily adheres to discrete technical protocols customized for each task, showcasing promising results\!~\cite{yang2019video,wang2021end,huang2022minvis,heo2022vita,wu2022seqformer,hu2020temporally,paul2021local,ji2023multispectral,sun2022coarse,kim2020video,weber2021step,woo2021learning,liang2023local,hui2023language,cheng2023segment,wang2015robust,wang2015saliency,wang2017super,lu2020learning,lu2020video}. Nevertheless, these approaches necessitate meticulous architectural designs for each unique task, thereby posing challenges in facilitating research endeavors devoting on one task to another. Recently, there have been efforts in shifting the above \textit{task-specific} paradigm to a \textit{general} solution that can be applied across multiple distinct tasks\!~\cite{li2022video,kim2022tubeformer,choudhuri2023context,athar2023tarvis,tubelink}.
However, one concern naturally arises that such a highly homogenized framework would overlook the diversity between tasks, potentially leading to suboptimal performance. 
For instance, the segmenting and tracking of objects like \textit{human} prioritize \textit{instance discrimination} in VIS but lean towards \textit{semantic recognition} in VSS. However, prior general approaches adopt exactly same query initialization, matching and space-time learning strategies\!~\cite{li2022video,kim2022tubeformer,tubelink}, lacking tailored differentiation within the algorithm design that caters to the specific properties of individual tasks.
\begin{figure}[t]
   \vspace{-5pt}
   \begin{center}
       \includegraphics[width=1.\linewidth]{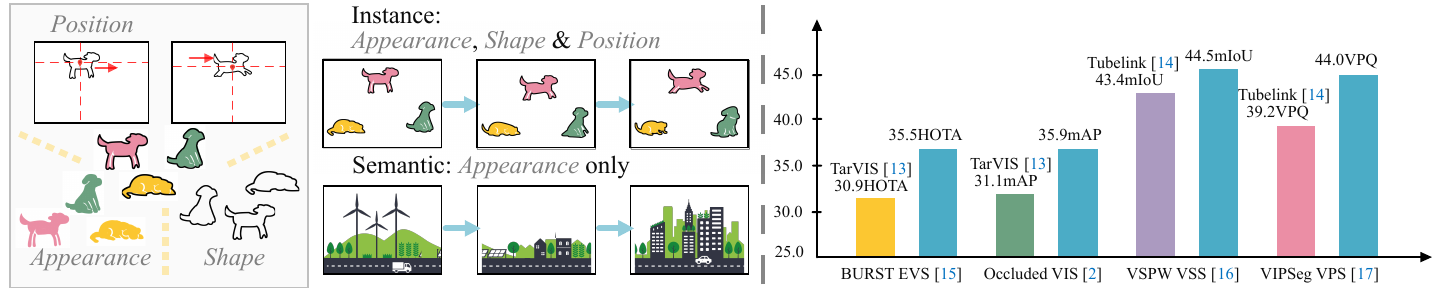}
       \put(-176.7,72.0){{\small (a)}}
       \put(-160.0,72.0){{\small (b)}}
       \end{center}
   \vspace{-18pt}
   \captionsetup{font=small}
   \caption{\small{(a) We render holistic modeling on segment targets by disentangling them into appearance, shape and position. {(b) By adjusting the involvement of the above} three factors into tracking and segmentation according to task requirement, \textsc{GvSeg} achieves remarkable improvement compared to prior top-leading general solutions.}
   } 
   \label{fig:1}
   \vspace{-12pt}
 \end{figure}

In this work, we present \textsc{GvSeg}, {a \textbf{g}eneral \textbf{v}ideo \textbf{seg}mentation framework to address EVS, VIS, VSS, and VPS that$_{\!}$ can seamlessly$_{\!}$ accommodate$_{\!}$ \textbf{\textit{task-oriented}} properties into the learning and inference process, while maintaining an \textbf{\textit{identical}} architectural design.} To achieve this, we rethink video segmentation in two aspects: \myhyperlink{Q1}{\ding{182}}\!~what are the key factors that constitute segment targets (\ie, \textit{instance}, \textit{thing}, and \textit{stuff}), and 
\myhyperlink{Q1}{\ding{183}}\!~how to leverage these key factors to build a unique sequential observation for each
specific task within a general model. 
To$_{\!}$ address$_{\!}$ \myhyperlink{Q1}{\ding{182}}, we delve deeply into the mechanism of how individuals can effectively discriminate moving instances or background stuff. The most intuitive answer in this regard is appearance, 
aligning with current video solutions where binary masks are classified solely based on visual representations (\ie, \textbf{appearance})\!~\cite{he2022inspro,heo2023generalized,huang2022minvis,qin2023coarse}. However, human perception extends beyond mere appearance\!~\cite{adelson2001seeing,loomis2002dissociation,wang2024visual}. 
For instance, we can also recognize moving entities such as cats in low-light conditions by referring to sketches (\ie, \textbf{shape}), and distinguish distinct instances on the basis of respective spatial locations (\ie, \textbf{position}), even in fast motion. Therefore, it is noteworthy that the instances to be segmented usually carry rich cues encompassing not only appearance but also position and shape characteristics.
In light of the analysis above, we could assert three significant observations that contribute to the resolution of \myhyperlink{Q1}{\ding{183}}: \textbf{First}, it becomes evident that current solutions downplay the importance of position and$_{\!}$ consistently$_{\!}$ ignore$_{\!}$ shape,$_{\!}$ in favor$_{\!}$ of$_{\!}$ solely$_{\!}$ appearance-based$_{\!}$ discrimination.$_{\!}$
To tackle this, we derive a \textit{shape-position descriptor} for each object, followed by encoding them into the cross-frame query matching process to enable the participation of three key factors
in discriminating corresponding instances across the entire video.
 \textbf{Second}, it is crucial to acknowledge that the engagement of appearance, position, and shape cues should be adjusted in accordance with the task requirements. In$_{\!}$ current$_{\!}$ {general}$_{\!}$ solutions,$_{\!}$ all$_{\!}$ queries are$_{\!}$ roughly$_{\!}$ initialized$_{\!}$ as$_{\!}$ empty and matched in the same manner. However, 
for semantic classes VSS and background \textit{stuff} in VPS, there is no instance discrimination and overly emphasize shape/location cues would harm the generalization of the model to various targets with the same semantics. Concerning this, we advocate for a tailored query initialization and object association strategies for each task by adjusting the relative contribution of three key elements. \textbf{Third}, owing to the absence of disentanglement on$_{\!}$ segment$_{\!}$ targets,$_{\!}$ the widely$_{\!}$ used$_{\!}$ temporal contrastive learning\!~\cite{huang2022minvis,li2022video,tubelink,wu2022defense} strategy for object association in$_{\!}$ current solutions is deemed suboptimal. Concretely, prior work empirically chooses objects in nearby frames as positive samples, remaining unaware of why excluding the same instance in distant frames.$_{\!}$ In fact,$_{\!}$ entities$_{\!}$ moving$_{\!}$ in long temporal range may display similar \textbf{appearance}, but undergo strong \textbf{shape} distortion, rendering them unsuitable as positive samples for instance discrimination.$_{\!}$ Therefore,$_{\!}$ we devise a task-oriented sampling strategy that caters$_{\!}$ to \textit{thing} and \textit{stuff},$_{\!}$ where$_{\!}$ instance examples are selectively$_{\!}$ sampled$_{\!}$ from$_{\!}$ the$_{\!}$ entire$_{\!}$ video$_{\!}$ by$_{\!}$ referring$_{\!}$ to$_{\!}$ shape$_{\!}$ similarity$_{\!}$ and$_{\!}$ location distance. This not only makes full use of the pre-defined \textit{shape-position descriptors}, but also recollects valuable samples that were arbitrarily discarded in prior work. In a similar spirit, the \textit{stuff} examples are gathered from the whole dataset which renders rich semantic description for each semantic class.    
Through an in-depth analysis of the essential elements that compose segmentation targets and subsequently derive task-oriented insights, our work exhibits several compelling facets: \textbf{First}, it not only recognizes but also effectively harnesses the unique nature of each task, enabling seamless accommodation of task-specific properties into segmentation models. \textbf{Second}, all of our designs are architecture-agnostic, preserving a uniform structural to efficiently address task diversity. \textbf{Third}, \textsc{GvSeg} substantially attains remarkable performance on each task. 
Notably, it surpasses existing general solutions by \textbf{4.6\%} HOTA on BURST\!~\cite{athar2023burst}, \textbf{1.3\%} AP on YouTube-VIS 2021\!~\cite{yang2019video}, \textbf{4.8\%} AP on Occluded-VIS\!~\cite{qi2022occluded}, \textbf{1.1\%} mIoU on VSPW\!~\cite{miao2021vspw}, \textbf{4.8\%} VPQ on VIPSeg\!~\cite{miao2022large}, establishing new SOTA.

\vspace{-2pt}
\section{Related Work} \label{sec:relatedwork}
\vspace{-2pt}
{\noindent\textbf{Exemplar-guided Video$_{\!}$ Segmentation$_{\!}$ (EVS).} Given the hint which can be mask, bounding box, or point at one video frame, EVS aims to propagate the mask-level predictions to subsequent frames\!~\cite{athar2023burst,athar2023tarvis}. Therefore, the standard video object segmentation (VOS) task can be viewed as a specific instance of EVS -- mask-guided video segmentation. Recent promising solutions for the mask-guided task mainly implemented in a \textit{matching-based} manner which classifies pixels in current frame according to the feature similarities of target objects in reference frames\!~\cite{yang2021collaborative,lu2020video,miao2020memory,wu2020memory,lu2020learning,wang2019zero,wang2019learning,lu2019see,wang2018semi,wang2015saliency,wang2019zero,seong2021hierarchical,mao2021joint,cheng2021rethinking,cheng2022xmem,li2022locality,park2022per,yu2022batman,zhang2023boosting,li2023unified}. To solve the bounding box and point-guided tasks, current solutions typically have to regress a pseudo ground-truth mask via pre-processing\!~\cite{athar2023burst,athar2023tarvis}. In contrast, \textsc{GvSeg} simply adapts various kinds of hints by initializing object queries from features within regions delineated by hints. 

\vspace{-1pt}
{\noindent\textbf{Video$_{\!}$ Instance$_{\!}$ Segmentation$_{\!}$ (VIS).} 
Extending beyond detecting and segmenting instances within images, VIS further engages in the active tracking of individual objects across video frames. According to the process of video sequences, existing solutions for VIS fall into three categories\!~\cite{heo2023generalized}: \textit{online}, \textit{semi-online}, and \textit{offline}. The \textit{online} methods take each frame as inputs and associate instances through hand-designed rules\!~\cite{yang2019video,cao2020sipmask,liu2021sg,yang2021crossover}, integrating learnable matching algorithms\!~\cite{han2022visolo,fang2021instances,zhu2022instance,li2021spatial,ke2021prototypical,lin2021video}, or deploying query matching frameworks\!~\cite{huang2022minvis,wu2022defense,he2022inspro,koner2023instanceformer,liu2023instmove,li2023mdqe}. The \textit{semi-online} solutions typically divide long videos into clips and model the representations of instances by leveraging rich spatio-temporal information\!~\cite{athar2020stem,wu2022efficient,yang2022temporally,kim2022tubeformer}. Conversely, \textit{offline} methods predict the instance sequence for an entire video in a single step\!~\cite{bertasius2020classifying,lin2021video,hwang2021video,wang2021end,wu2022seqformer,heo2022vita} which require a growing amount of GPU memory as the video length extends, limiting their application in real-world scenarios.

\vspace{-1pt}
\noindent\textbf{Video$_{\!}$ Semantic$_{\!}$ Segmentation$_{\!}$ (VSS).}  
{Building$_{\!}$ upon$_{\!}$ the principle$_{\!}$ of$_{\!}$ semantic$_{\!}$ segmentation\!~\cite{wang2021exploring,zhou2022rethinking,li2023logicseg,chen2023pipa,li2022deep,li2023semantic,chen2023transferring,zhou2024cross}},$_{\!}$ VSS$_{\!}$ extends$_{\!}$ this$_{\!}$ concept$_{\!}$ to$_{\!}$ video$_{\!}$ sequences,$_{\!}$ so$_{\!}$ as$_{\!}$ to$_{\!}$ capture the evolution of scenes$_{\!}$ and$_{\!}$ objects$_{\!}$ over$_{\!}$ time.$_{\!}$} Existing$_{\!}$ solutions can generally be classified into two main paradigms.$_{\!}$ The \textit{motion-based} approaches\!~\cite{xu2018dynamic,mahasseni2017budget,nilsson2018semantic,jain2019accel,liu2020efficient}$_{\!}$ employ$_{\!}$ optical$_{\!}$ flow$_{\!}$ to$_{\!}$ model dynamic$_{\!}$ scenes.$_{\!}$ Though workable$_{\!}$ in certain scenarios,$_{\!}$ they rely heavily$_{\!}$ on the accuracy of$_{\!}$ flow maps and are prone to error accumulation\!~\cite{zhou2022survey}.$_{\!}$ On$_{\!}$ the$_{\!}$ other hand, the \textit{attention-based} methods take advantage of the attention mechanism\!~\cite{paul2021local,ji2023multispectral,sun2022coarse} or Transformer\!~\cite{li2021video,sun2022mining}$_{\!}$ to aggregate temporal cues.$_{\!}$ This contributes to improved coherence among predictions of individual frames.  
\vspace{-1pt}
\noindent\textbf{Video$_{\!}$ Panoptic$_{\!}$ Segmentation$_{\!}$ (VPS).} With the emergence of seminal work\!~\cite{kim2020video},$_{\!}$ there has been a research trend\!~\cite{woo2021learning,qiao2021vip,kreuzberg20224d,zhou2022slot,yuan2022polyphonicformer,he2023towards,shin2023video} dedicated to unifying video instance and semantic segmentation. 
Though$_{\!}$ showing$_{\!}$ the$_{\!}$ promise$_{\!}$ of$_{\!}$ general video segmentation, the early work\!~\cite{woo2021learning,qiao2021vip,kreuzberg20224d} utilizes task-specific heads to handle instance and semantic segmentation separately, and assembles the panoptic predictions through post-processing. Recent algorithms typically leverage unified queries for the detection and tracking of both \textit{thing} and \textit{stuff} objects\!~\cite{zhou2022slot,yuan2022polyphonicformer,he2023towards,shin2023video}. However, they$_{\!}$ demonstrate$_{\!}$ sub-optimal$_{\!}$ performance compared$_{\!}$ to$_{\!}$ task-specified solutions, emphasizing the urgency for the development of more powerful solutions.

\noindent\textbf{General$_{\!}$ Video$_{\!}$ Segmentation (GVS).}$_{\!}$ In order to address the limitations of task-specific models that lack the flexibility to generalize across different tasks and result in redundant research efforts, GVS aims at an all-inclusive solution for multiple video segmentation tasks.
A limited number of studies\!~\cite{cheng2022masked,li2022video,choudhuri2023context,kim2022tubeformer,athar2023tarvis,tubelink,zhang2023dvis}$_{\!}$ have ventured in this direction. However, \cite{cheng2022masked,li2022video,kim2022tubeformer} exhibits inferior performance compared to dedicated, task-specific methods. \cite{athar2023tarvis} achieves remarkable results but requires extensive pre-training on various large-scale, pixel-level annotated datasets. Inspired$_{\!}$ by$_{\!}$ these pioneers, \textsc{GvSeg} \textbf{i)} delves deeper into the segment targets across tasks, offering a disentanglement and modeling for them, \textbf{ii)} harnesses insights gained from \textbf{i)} to adapt task-oriented property without any modification to network architecture or training objectives, and \textbf{iii)} contributes to a {robust} solution that outperforms all existing specialized/general models.
\vspace{-1pt}
\noindent\textbf{Query-Based$_{\!}$ Segmentation.} Image segmentation has witnessed substantial progress with top-performing approaches primarily falling into the \textit{query-based} paradigm. Such paradigm directly models targets by introducing a set of learnable embeddings as queries to search for objects of interest and subsequently decode masks from image features. Inspired by DETR\!~\cite{carion2020end}, the latest research\!~\cite{cheng2022masked,cheng2021per,wang2022learning,ding2024clustering,liang2023clustseg} takes this paradigm a step further by harnessing the Transformer architecture. This trend also spills over into video segmentation with recent solutions\!~\cite{cheng2022masked,li2022video,kim2022tubeformer,athar2023tarvis,tubelink} all building upon their image segmentation counterparts.
In contrast to prior work that focused solely on object appearance, \textsc{GvSeg} provides a holistic modeling of targets by encoding the relative position and shape cues into queries. This is particularly valuable for the tracking of instance objects. As a result, the query matching process can harness appearance, shape, and position information, enhancing object association across frames.

\section{Methodology} \label{sec:methodology}
\vspace{-2pt}
\noindent{\textbf{Problem Statement.} Video segmentation seeks to partition a video clip $V\!\in\!\mathbb{R}^{THW\times3}$ containing $T$ frames of size $H\!\times\!W$ into $K$ non-overlap tubes linked along the time axis:
\vspace{-4pt}
\begin{equation}
\begin{aligned}\label{eq:1}
\{Y_k\}_{k=1}^{K}=\{(M_k, c_k)\}_{k=1}^{K},
\end{aligned}
\vspace{-2pt} 
\end{equation}
where each tube mask $M_k\!\in\!\{0, 1\}^{T\!\times\!H\!\times\!W\!}$ is labeled with a category $c_k\in\!\{1, \cdots, C\}$. The value of $K$ varies across tasks:$_{\!}$ in VSS,$_{\!}$ it is consistent with the number of predefined semantic categories; in EVS and VIS,$_{\!}$ it$_{\!}$ is adjusted in response to the instance count; and in VPS,$_{\!}$ it is the sum of \textit{stuff} categories and \textit{thing} entities.}
 
\noindent\textbf{Tracking by Query Matching.} Inspired by the success of \textit{query-based} object detectors, \cite{huang2022minvis,wu2022defense,li2022video} propose to associate instances based on the query embeddings. Specifically, given a set of $N$  randomly initialized queries $\{\bm{q}_n^t\}_{n=1}^N$, we can derive the object-centric representation $\{\hat{\bm{q}}_n^{t}\}_{n=1}^N$ for frame $V^t$ by:
\vspace{-2pt}
\begin{equation}
\begin{aligned}\label{eq:2}
\{\hat{\bm{q}}_n^{t}\}_{n=1}^N=\mathcal{D}(\mathcal{E}({V^t}), \{\bm{q}_n^{t}\}^N_{n=1}),
\end{aligned}
\vspace{-2pt}
\end{equation}
where $\mathcal{E}$ and $\mathcal{D}$ are the Transformer encoder and decoder. Here $\hat{\bm{q}}^t_n$ refines rich appearance representation for a specific object.$_{\!}$ The tracking is done by applying Hungarian Matching on the affinity matrix $\mathcal{S}_{ij}{\!} ={\!} \texttt{cosine}(\hat{\bm{q}}_i^{t}, \hat{\bm{q}}_j^{t+1})$ computed between $\hat{\bm{q}}_i^{t}$ and $\hat{\bm{q}}_j^{t+1}$ of two successive frame $V^t$ and $V^{t+1}$. As such, instances exhibiting identical attributes across the video sequence are linked automatically.

\subsection{\textbf{\textsc{GvSeg}}: Task-Oriented Property Accommodation Framework}\label{sec:3.2}
\vspace{-2pt}
\textsc{GvSeg} seeks to {advance general video segmentation} through controllable emphasis on instance discrimination and semantic comprehension according to task requirements.
Concretely, we first devise a new
shape-position descriptor to accurately reveal the shape and location of targets. 
Then, by adjusting the engagement of above shape-position descriptor during cross-frame query matching,
we could realize controllable association for instance and background stuff, respectively. 
Finally, we give an analysis on the limitation of current temporal contrastive learning and devise a task-oriented sampling strategy to tackle encountered issues.

\begin{figure}[t] 
   \centering
   \vspace{-5pt}
       \vspace{+5pt}
       \centering
       \includegraphics[width=\linewidth]{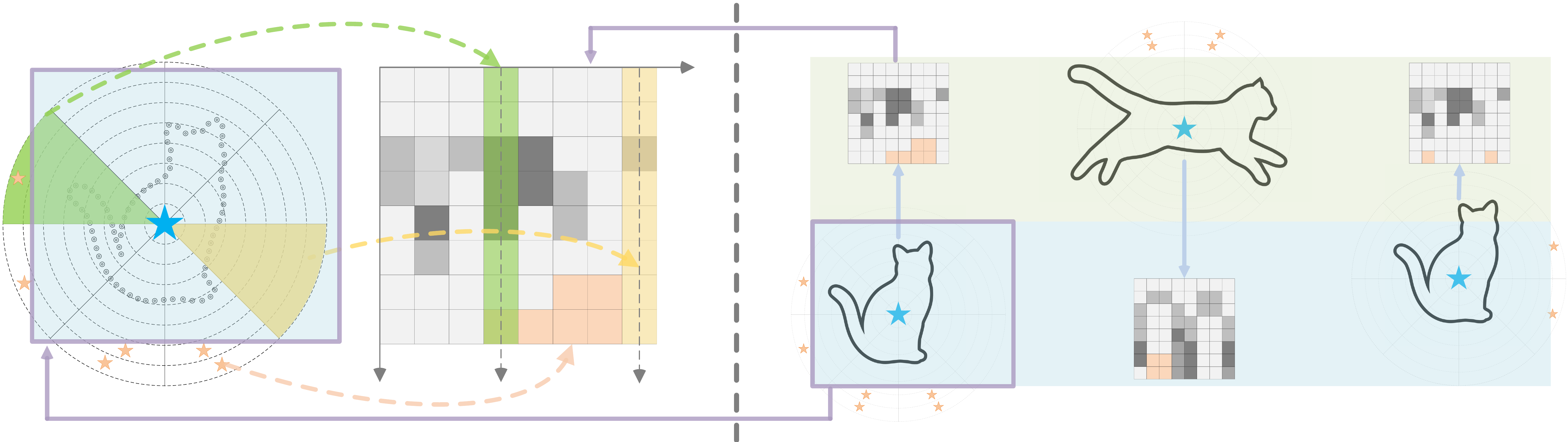}
       \put(-307.3,45){{\tiny $p_o$}}
       \put(-83.5,67.5){{\tiny $A_{p_o}$}}
       \put(-155.5,21.6){{\tiny $B_{p_o}$}}
       \put(-30.4,29.4){{\tiny $C_{p_o}$}}
       \put(-195.4,80.7){{ $u$}}
       \put(-267.7,7.7){{ $v$}} 
   \hfill%
   \vspace{-7pt} 
   \captionsetup{font=small}
   \caption{\small{Illustration of \textbf{shape-position descriptor}} (\S\ref{sec:3.2}).}
   \label{fig:2.2}
   \vspace{-15pt}
 \end{figure}
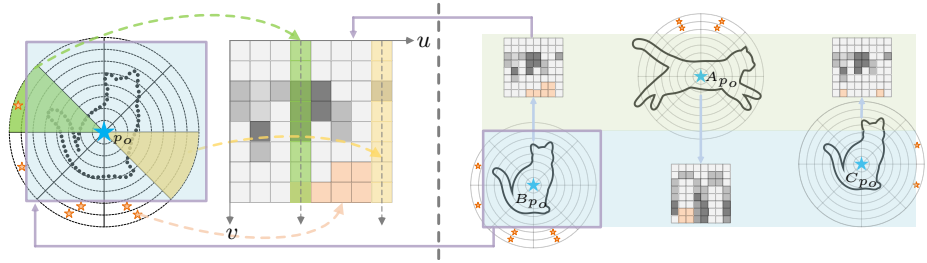

\noindent\textbf{Shape-Position Descriptor.}
{Inspired by shape context\!~\cite{belongie2002shape}, a \text{shape-position} descriptor }is constructed to represent the spatial distribution and shape of target objects.
{First, it describes shape cues by encoding the relative geometric relationships of points in object contours relative to the object center.}
As shown in Fig.\!~\ref{fig:2.2}, given the contour $G\!\in\!\{0,1\}^{H\!\times\!W}$ of a target object which can be easily derived from masks, a set $P$ with $M$ anchor points (\ie, \includegraphics[scale=0.04,valign=c]{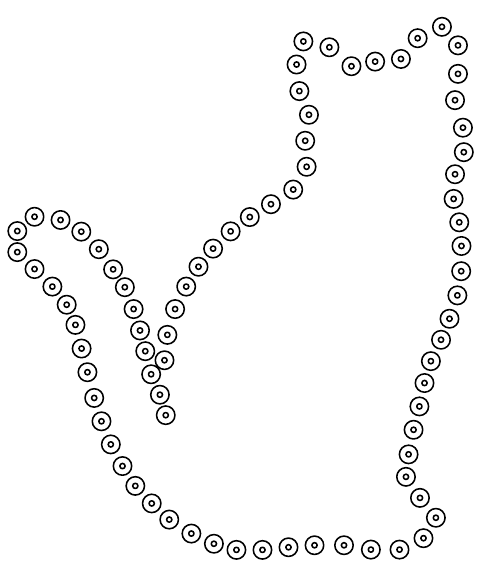}) are evenly sampled: 
\vspace{-2pt}
\begin{equation} 
\begin{aligned}\label{eq:3}
\mathcal{P} = \{p_m = (x, y) \, | \, G(x, y) = 1, \, 1 \leq m \leq M\}.
\end{aligned}
\vspace{-2pt}
\end{equation} 
Above anchor points are transformed into polar coordinates with the {central point} $p_o$ of targets (\ie, \includegraphics[scale=0.18,valign=t]{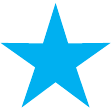}) as the reference point.
The polar coordinate is a histogram divided into a grid of $u\!\times\!v$ bins with $u$ angle divisions and $v$ radius divisions.
Next we calculate the number of anchor points falling within each bin:
\vspace{-2pt}
\begin{equation}\label{eq:4}
    \!\!\!\bm{H}_{i,j}\! =\! \sum_{m=1}^{M} \left\{
    \begin{array}{ll}
        \frac{1}{\sqrt{d_\text{model}}} &  \ \ \ \text{if} \ \ \ | \theta_m - \hat\theta_i | \leq \frac{\Delta\theta}{2} \text{\ \ and\ \ } | r_m - \hat r_j |\! \leq\! \frac{\Delta r}{2}\\
        \ \ \ 0 & \ \ \ \text{otherwise}
    \end{array}
\right\}, 
\vspace{-2pt}
\end{equation}
where $\Delta \theta$, $\Delta r$, and $(\hat\theta_i, \hat r_j)$ are the angle span, radius span, and center point of each bin, $(\theta_m, r_m)$ is the polar coordinate of anchor point $p_m$, $d_\text{model}$ is the embedding dimension of model. As such, $\bm{H}$ expresses the spatial configuration of contour $G$ relative to {center point} (\ie, $p_o$) 
in a compact and robust way.  
As depicted in Fig.\!~\ref{fig:2.2}, 
instances with different shapes (\ie, target $A$ and $B$) present varying distributions of $\bm{H}$ which 
demonstrates the capability to encode the shape cues of target objects.
Moreover, we equip $\bm{H}$ with the ability to account for the relative spatial location of target objects by setting $\bm{H}_{i,j}=-1/\sqrt{d_\text{model}}$ if the center point of a bin (\ie, \includegraphics[scale=0.38,valign=t]{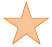}) falls outside of masks. Therefore, instances with similar shapes but different locations (\ie, target $B$ and $C$) would yield similar distribution of positive values, but distinct distribution of negative values, effectively evolving above \text{shape} descriptor into a \textbf{shape-position} descriptor. 

\noindent\textbf{Shape- and Position-Aware (SPA) Query Matching.}
Given the above analysis, a set of shape-position descriptors $\{\bm{H}_k\}_{k=1}^K$ could be derived from each object $k$ within the mask. We then aim to facilitate the awareness of shape-position cues for object association between frames, by integrating such descriptors into the query matching process. 
To achieve this, as shown in Fig.\!~\ref{fig:2.1} (c), we draw inspiration from the absolute position encoding (APE) which is widely adopted in Transformer\!~\cite{vaswani2017attention}. Specifically, during mask decoding, $N$ query embeddings $\{\bm{q}_n\}_{n=1}^N$ is interacting with the backbone feature $\bm{F}$ to retrieve object-centric feature in each decoder layer by:
\vspace{-3pt}
\begin{equation} 
\begin{aligned}\label{eq:5}
\bm{q}^{l} = \texttt{CrossAttn}(\bm{q}^{l-1}, \bm{F}), \ \ \  \bm{q}^{l} = \texttt{SelfAttn}(\bm{q}^{l}, \bm{q}^{l})
\end{aligned} 
\vspace{-2pt}
\end{equation}
Where $l$ is the layer index.
Typically, a Hungarian Matching matrix $\mathbbm{1}^{l}\!\in\!\{0,1\}^{N\!\times\!K}$ between $N$ predictions generated from query embeddings and $K$ ground truth objects can be derived from each decoding layer. 
Following the principle of APE, where the position encodings $\mathbf{P}$ is integrated into $\bm{q}$: $\bm{q}\leftarrow \bm{q}+\mathbf{P}$, we assign $\{\bm{H}_k\}_{k=1}^K$ to $K$ elements in $\bm{q}$ that corresponds to the object described in ground truth by referring to $\mathbbm{1}^{l-1}$ produced from prior decoding layer: $\bm{{q}}^l \leftarrow \bm{q}^l + \mathbbm{1}^{l-1} \cdot \bm{H}$ before conducting \texttt{SelfAttn}.
Note the $K$ elements in $\{\bm{H}_k\}_{k=1}^K$ are flattened and bilinearly interpolated to size $d_\text{model}$, and then stacked together to get $\bm{H}\in\mathbb{R}^{K\!\times d_\text{model}}$.
In this way, the query embeddings can \textbf{i)} well attend to and discriminate corresponding objects by injecting the descriptors into \texttt{SelfAttn}, and \textbf{ii)} be aware to shape-position cues after mask decoding (\ie, $\hat{\bm{q}}$ in Eq.\!~\ref{eq:2}). To further reinforce the consideration to shape and position of targets in $\hat{\bm{q}}$, we compile $\bm{H}$ into the affinity-based query matching between two adjacent frames:
 \vspace{-4pt}
\begin{equation}
\begin{aligned}\label{eq:7}
\mathcal{S}_{ij}{\!} ={\!} \texttt{cosine}(\hat{\bm{q}}_i^{t}+\bm{H}_i^{t}, \hat{\bm{q}}_j^{t+1}+\bm{H}_j^{t+1}).
\end{aligned} 
\vspace{-4pt} 
\end{equation} 
As such, each query embedding is seamlessly incorporated with the unique attributes of corresponding objects, 
thereby {endowing them with a heightened sensitivity to specific targets when matching with other frames afterward. }

\noindent\textbf{Task-Oriented Query Initialization \& Object Association.} To orient the model towards specific tasks, existing$_{\!}$ work usually employs dedicated queries (\ie, \textit{stuff}/\textit{thing} query) for semantic/instance$_{\!}$ segmentation\!~\cite{li2022panoptic,yuan2022polyphonicformer}, and process them parallel by modifying$_{\!}$ the model into a two-path architecture.$_{\!}$ In contrast, \textsc{GvSeg} smartly addresses this$_{\!}$ challenge by$_{\!}$ dynamically adjusting the involvement of three key constitutes, \ie, \textbf{appearance}, \textbf{shape}, and \textbf{position} within the query initialization (\ie, Fig.\!~\ref{fig:2.1} (a)) and object association (\ie, Fig.\!~\ref{fig:2.1} (b)) according to task requirements.\\
\noindent$\bullet$\ \textbf{EVS} underscores the utilization of given hints to guide the segmentation of subsequent frames.$_{\!}$ To flexibly unleash the potential of different kinds of hints under the \textit{track by query matching} paradigm, we propose to initialize the query embeddings$_{\!}$ from$_{\!}$ backbone features sampled within hinted regions. Specifically, for the point-guided task which provides a single point $p_k=(x,y)$ to indicate the target$_{\!}$ object,$_{\!}$ the$_{\!}$ backbone$_{\!}$ feature$_{\!}$ at$_{\!}$ corresponding$_{\!}$ location$_{\!}$ can$_{\!}$ be$_{\!}$ sample{d} by:
\vspace{-3pt} 
\begin{equation}
\begin{aligned}\label{eq:8} 
    \bm{f}_k =\texttt{sample}({\bm{F},\ p_k}),
\end{aligned}
\vspace{-3pt} 
\end{equation}
where the implementation of $\texttt{sample}$ follows PointRent\!~\cite{kirillov2020pointrend}. Then, the query embedding is initialized with $f_k$: $\bar{\bm{q}}_k =\texttt{FFN}({\bm{f}_k})$ to fulfill the guidance ability of given exemplars where $\texttt{FFN}$ is a feed-forward network. For the mask and box guided tasks, we sample multiple $f_k$ and average them to get the feature that comprehensively describes target objects. 
Finally, SPA query matching is applied to enhance instance discrimination during the object association between frames.

\noindent$\bullet$\ \textbf{VIS} emphasizes the tracking of instances which usually exhibits unique attributes for discrimination. To encode these instance-specific properties (\eg, location, appearance) into query embeddings, we follow\!~\cite{wang2020solov2} to initialize $\bm{q}\!\in$ $\mathbb{R}^{N\!\times\!D}$ from the backbone features. Concretely, we partition the backbone features into $S\times S$ grids and flatten them, resulting in $\{\bm{F}_{i}\}^{S\times S}_{i=1}$. We then randomly select $N$ elements from this set for the initialization of queries and obtain $\{\bar{\bm{q}}_{i}\}^{N}_{i=1}$:
\vspace{-3pt} 
\begin{equation}
\begin{aligned}\label{eq:8} 
    [\bar{\bm{q}}_0;\cdots\!;\bar{\bm{q}}_{N}] =\texttt{FFN}({\bm{F}}).
\end{aligned}
\vspace{-3pt}
\end{equation}
As$_{\!}$ such,$_{\!}$ queries could involve appearance and location cues for diverse instances present in the frame. Similarly to EVS, we apply SPA query matching for object association to enable more precise instance discrimination across the entire video.

\noindent$\bullet$\ \textbf{VSS} prioritizes semantic understanding of each class. Therefore, to enhance the$_{\!}$ thorough grasp$_{\!}$ of$_{\!}$ semantics, we continuously collect the query embeddings corresponding to each semantic class during training.  
More precisely, given $N$ queries $\bm{q}\!\in$ $\mathbb{R}^{N\!\times\!D}$, we gather $K$ entities from them based on the bipartite matching results $\mathbbm{1}\!\in\!\{0,1\}^{K\!\times\!N}$ between predictions generated from $\bm{q}$ and ground truth:
\vspace{-4pt}  
\begin{equation} 
\begin{aligned}\label{eq:9}
    \bar{\bm{q}} = \mathbbm{1}\odot\bm{q} \in \mathbb{R}^{K\!\times\!D}.
    \end{aligned} 
\vspace{-2pt}
\end{equation}
Here $\bar{\bm{q}}$ encodes the semantic-specific properties for each class, and we momentously update it in each training step to approximate the global representation of semantic classes over the entire dataset. During inference, we initialize object queries for each frame from $\bar{\bm{q}}$. 
Note we do not apply SPA query matching for VSS, as shape and location cues would harm semantic-level tracking.
 
\noindent$\bullet$\ \textbf{VPS}$_{\!}$ integrates$_{\!}$ both$_{\!}$ instance-discrimination$_{\!}$ for foreground \textit{thing} classes and semantic interpretation for background \textit{stuff} categories.  We thus combine the query initialization and association strategies used in VIS and VSS, to facilitate the effective recognition and tracking for \textit{thing} and \textit{stuff} classes, respectively.


\begin{figure*}[t]
  \begin{center}
      \includegraphics[width=1.\linewidth]{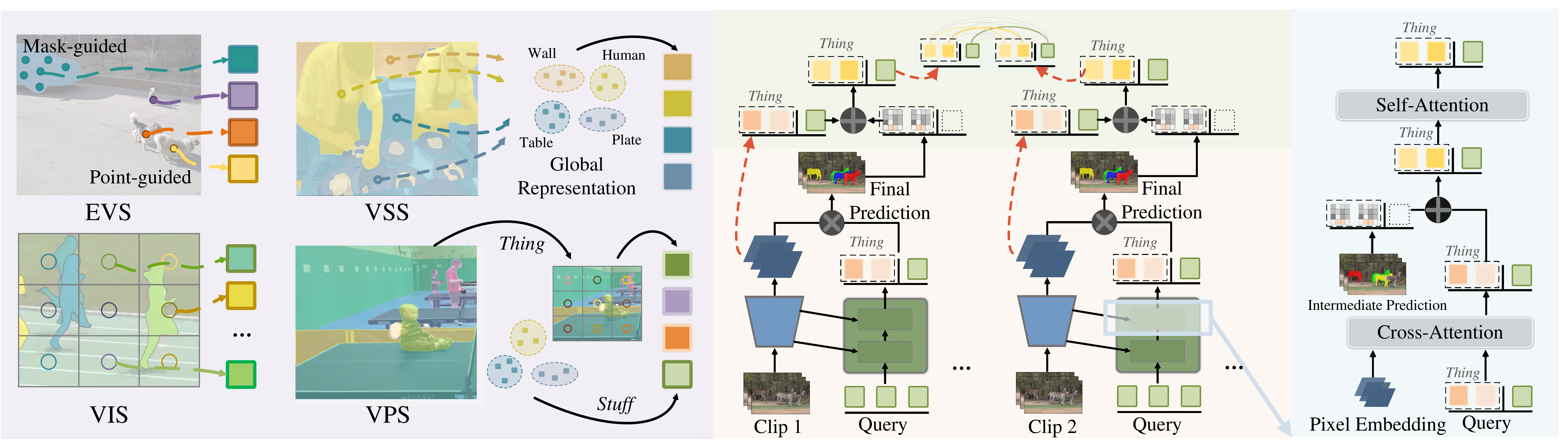}
      \put(-199.7,90.8){{\scriptsize (a)}}
      \put(-188.7,90.8){{\scriptsize (b)}}
      \put(-58.79, 90.8){{\scriptsize (c)}}
      \put(-47.79, 54.7){{\tiny ${\bm{H}}$}} 
      \put(-88.79, 63.7){{\tiny ${\bm{H}}$}}
      \put(-148.79,63.7){{\tiny ${\bm{H}}$}}
      \end{center}
  \vspace{-18pt}
  \captionsetup{font=small}
  \caption{\small{(a) {Task-oriented queries initialization}.  (b) Task-oriented object association tailored \wrt \textit{thing} and \textit{stuff} objects. (c) Shape- and position-aware query matching.}
  }
  \label{fig:2.1} 
  \vspace{-10pt}
\end{figure*}
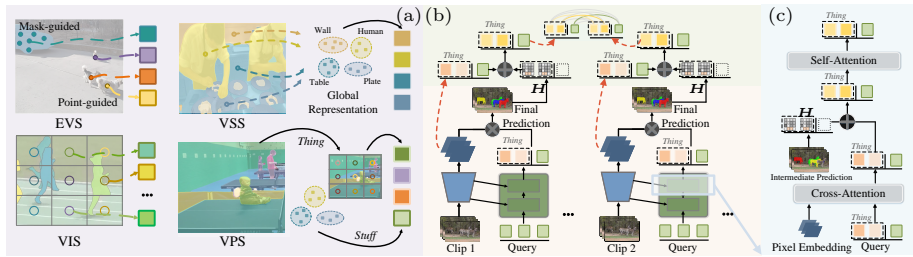


\noindent\textbf{Task-Oriented Temporal Contrastive Learning.} 
 The performance of current \textit{track by query matching-based} solutions depends significantly on the temporal contrastive learning (TCL) between frames. Given a key frame, prior methods\!~\cite{li2022video,wu2022defense,tubelink} typically select reference frames from the temporal neighborhood, while ignoring all other frames. This leads to limited positive/negative samples for effective contrastive learning which relies on a substantial quantity of samples to achieve optimal performance. To maximize the usage of these discarded samples, we devise a smart sampling strategy that caters to individual tasks and addresses the challenge of accurately distinguishing the positive ones from them (\ie, Fig.\!~\ref{fig:4}).
 Specifically, for tasks leaning towards {instance discrimination} (\ie, VIS, EVS and \textit{thing} in VPS),
 it is essential to note that not all identical instances in the same video are suitable as positive samples. This is due to the strong variations in shape and spatial location among instances, which can disrupt the local consistency between the same instance at nearby frames that usually manifest similar shape and position. To tackle this, in contrast to existing work arbitrarily discards samples in distant frames, we sample examples across the whole video by measuring the shape and location similarity.
The variation of shape-position descriptors (\ie, $\Delta H$) belonging to the same instance but at frame $V^t$ and $V^{t+n}$ is computed via:
\vspace{-4pt}
\begin{equation}
\begin{aligned}\label{eq:10}
\!\!\!\!\Delta H = \frac{\lVert \bm{H}^{t+n}-\bm{H}^{t} \lVert_2}{\lVert \bm{H}^t \lVert_2}.
    \end{aligned}
\vspace{-2pt} 
\end{equation}
We set a threshold $\tau = 0.2$ and consider the query embedding associated with $\bm{H}^{t+n}$ as a positive example if $\Delta H$ is smaller than $\tau$; otherwise, it is deemed negative. 
As such, we involve distant frames into the reference set which enriches the diversity of samples and 
bolsters the robustness of TCL. 
On the other hand, for VSS and background \textit{stuff} classes in VPS, samples are relaxed to select from the whole training set, as$_{\!}$ larger mount of entities with diverse appearance, shape, and location will$_{\!}$ improve$_{\!}$ the$_{\!}$ grasp$_{\!}$ of$_{\!}$ semantics. To implement this, we maintain a first-in-first-out queue $\mathcal{Q}$ that contains $N_\mathcal{Q}$ queries for each pre-defined semantic class. Elements in $\mathcal{Q}$ will engage in TCL and be updated with new samples at each training step.
We set $N_\mathcal{Q}$ to a relatively small number (\eg, 100), which incurs negotiable cost in training time but considerable improvement in performance. 

\begin{figure}[t] 
  \centering
  \vspace{-0pt}
      \vspace{+20pt}
      \centering
      \includegraphics[width=\linewidth]{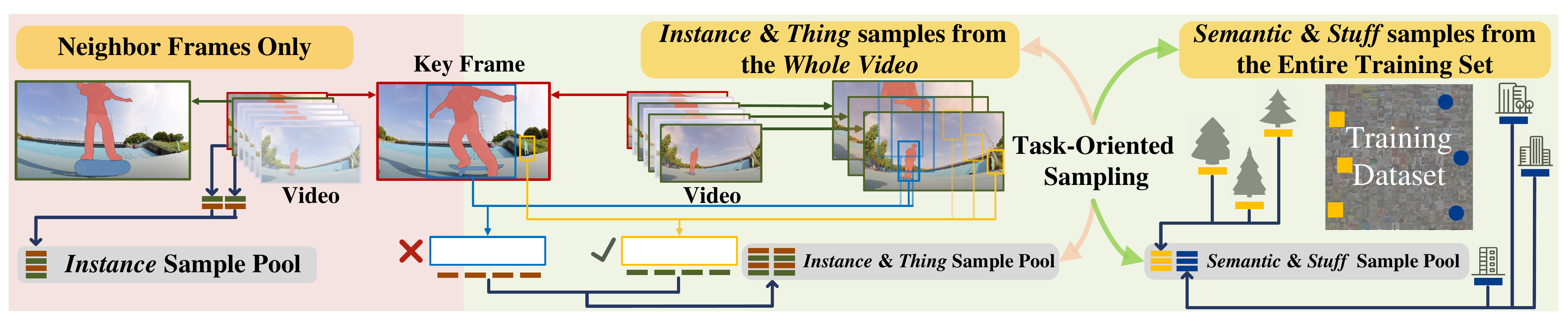}
      \put(-207.0,14.9){{\tiny $\Delta\! H\! <\! \tau$}}
      \put(-249.8,14.9){{\tiny $\Delta\! H\! >\! \tau$}}
  \hfill%
  \vspace{-5pt} 
  \captionsetup{font=small}
  \caption{\small{Illustration of \textbf{task-oriented temporal contrastive learning} (\S\ref{sec:3.2}). Prior work considers solely \textit{instance} objects, and samples are restricted within neighbor frames. In \textsc{UvSeg}, \textit{instance} \& \textit{thing} samples are collected from the whole video according to shape and location similarity, while \textit{semantic} \& \textit{stuff} samples are gathered from the entire training set to capture diver shapes and appearances of each semantic class.}}
  \label{fig:4}
  \vspace{-10pt}
\end{figure}

\subsection{Implementation Details}\label{sec:3.3}
\vspace{-2pt}
\noindent\textbf{Network$_{\!}$ Configuration.}$_{\!}$ \textsc{GvSeg} is a semi-online video segmentation framework built upon the \textit{tracking by query matching} paradigm\!~\cite{huang2022minvis}. It comprises an \textit{image-level segmenter} to extract frame-level queries, and an object associator to match query embeddings across frames. The \textit{image-level segmenter} is implemented as  Mask2Former\!~\cite{cheng2022masked} with both ResNet-50\!~\cite{he2016deep} and Swin-L\!~\cite{liu2021swin} as the backbone. Given the most recent work typically adopts clip-level inputs for richer temporal cues\!~\cite{heo2023generalized,tubelink,heo2022vita}, in alignment with this trend, \textsc{GvSeg} takes a clip containing three frames as input each time. The size of points set $\mathcal{P}$ derived from object contour is fixed to 200  to make the shape-position descriptor effectively characterize objects of varying scales. We employ $u=36$ angle divisions and $v=12$ radius divisions to capture point distribution in finer granularity.

\noindent\textbf{Training.$_{\!}$} 
Following the standard protocols~\cite{kim2022tubeformer,cheng2021mask2former,heo2022vita,tubelink,cheng2021rethinking} in video segmentation, the maximum training iteration is set to $10$K for OVIS/VSPW/VIPSeg/KITTI and $15$K for YouTube-VOS$_{18}$/YouTube-VIS$_{21}$ with a mini-batch size of 16. The AdamW optimizer with initial learning rate 0.001 is adopted. The learning rate is scheduled following a step policy, decayed by a factor of $10$ at $7$K/$11$K for $10$K/$15$K total training steps, respectively. Following existing solutions~\cite{wu2022defense,wu2022seqformer,heo2023generalized,he2022inspro}, we generate pseudo videos from MS COCO\!~\cite{lin2014microsoft} as training samples for YouTube-VOS$_{18}$/YouTube-VIS$_{21}$ while no additional data is used for other benchmarks.  We use standard data augmentations, \ie, flipping, random scaling and cropping.  The \textit{frame segmenter} is initialized with weights pre-trained on MS COCO.

\noindent\textbf{Testing.$_{\!}$} The evaluation process follows existing work\!~\cite{tubelink,athar2023tarvis,UNINEXT,cheng2021rethinking} and adopts no test-time augmentation to ensure a fair comparison. For YouTube-VOS$_{18}$/YouTube-VIS$_{21}$, videos are resized to 360p/480p for ResNet/Swin backbones. For OVIS/VSPW/VIPSeg/KITTI/BURST, videos are tested at a resolution of 720p.

\noindent\textbf{Reproducibility.} \textsc{GvSeg} is implemented in PyTorch and trained on eight Tesla A40 GPUs. The testing is conducted on one Tesla A40 GPU.

\section{Experiment}
\label{sec:experiments}
  \vspace{-3pt}

  \begin{table*}[t]
   \caption{
      Quantitative results for VPS on VIPSeg\!~\cite{miao2022large} and KITTI-STEP\!~\cite{weber2021step} (\S\ref{sec:VPS}),  and VSS on VSPW\!~\cite{miao2021vspw} (\S\ref{sec:VSS}).
    }
    \vspace{-6pt}
    \resizebox{\linewidth}{!}{
          \setlength\tabcolsep{1.3pt}
          \renewcommand\arraystretch{0.99}
          \begin{tabular}{r|c|c|cccc|cccc|ccc}
          \thickhline
           \multirow{2}{*}{Method} & \multirow{2}{*}{Backbone} & {General} & \multicolumn{4}{c|}{VIPSeg \texttt{val}}  & \multicolumn{4}{c|}{KITTI-STEP \texttt{val}}& \multicolumn{3}{c}{VSPW \texttt{val}}  \\
    \cline{4-14} 
          &                   & Solution & VPQ & {VPQ}$^{\text{Th}}$ & {VPQ}$^{\text{St}}$ & STQ &VPQ & STQ & AQ & SQ  & mIoU & mVC$_\text{8}$ & mVC$_\text{16}$ \\
    \hline
    \color{mygray2}VPSNet~\cite{kim2020video}     & \color{mygray2}R-50 & \color{mygray2}\xmark & \color{mygray2}14.0 & \color{mygray2}14.0 & \color{mygray2}14.2 & \color{mygray2}20.8 & \color{mygray2}0.43 & \color{mygray2}0.56 & \color{mygray2}0.52 & \color{mygray2}0.61 & - & - & -  \\
    \color{mygray2}Mask-Prop~\cite{kim2020video}  & \color{mygray2}R-50 & \color{mygray2}\xmark & \color{mygray2}- & \color{mygray2}- & \color{mygray2}- & \color{mygray2}- & \color{mygray2}- & \color{mygray2}0.67 & \color{mygray2}0.63 & \color{mygray2}0.71 & \color{mygray2}- & \color{mygray2}- & \color{mygray2}- \\
    \color{mygray2}MotionLab~\cite{weber2021step}  & \color{mygray2}R-50 & \color{mygray2}\xmark & \color{mygray2}- & \color{mygray2}- & \color{mygray2}- & \color{mygray2}- & \color{mygray2}0.40 & \color{mygray2}0.58 & \color{mygray2}0.51 & \color{mygray2}0.67 & \color{mygray2}- & \color{mygray2}- & \color{mygray2}- \\
    \color{mygray2}SiamTrack~\cite{woo2021learning} & \color{mygray2}R-50 & \color{mygray2}\xmark & \color{mygray2}17.2 & \color{mygray2}17.3 & \color{mygray2}17.3 & \color{mygray2}21.1 & \color{mygray2}- & \color{mygray2}- & \color{mygray2}- & \color{mygray2}- & \color{mygray2}- & \color{mygray2}- & \color{mygray2}- \\
    \color{mygray2}TCB~\cite{miao2021vspw} & \color{mygray2}R-101 & \color{mygray2}\xmark & \color{mygray2}- & \color{mygray2}- & \color{mygray2}- & \color{mygray2}- & \color{mygray2}- & \color{mygray2}- & \color{mygray2}- & \color{mygray2}- & \color{mygray2}37.5 & \color{mygray2}86.9 & \color{mygray2}82.1  \\
    \color{mygray2}DVIS~\cite{zhang2023dvis} & \color{mygray2}R-50 & \color{mygray2}\xmark & \color{mygray2}43.2 & \color{mygray2}43.6 & \color{mygray2}42.8 & \color{mygray2}42.8 & \color{mygray2}- & \color{mygray2}- & \color{mygray2}- & \color{mygray2}- & \color{mygray2}- & \color{mygray2}- & \color{mygray2}-  \\
      Mask2Former~\cite{cheng2022masked} & R-50 & \cmark & - & - & - & - & - & - & - & - & 38.4 & 87.5 & 82.5 \\
    TubeFormer~\cite{kim2022tubeformer} & R-50 & \cmark & 26.9 & - & - & 38.6 & 0.51 & 0.70 & 0.64 & 0.76 & - & - & -    \\
    Video K-Net~\cite{li2022video} & R-50 & \cmark & 26.1 & - & - & 31.5 & 0.46 & 0.71 & 0.70 & 0.71 & 37.9 & 87.0 & 82.1 \\
    TarVIS~\cite{athar2023tarvis} & R-50 & \cmark & 33.5 & 39.2 & 28.5 & 43.1 & - & 0.70 & 0.70 & 0.69 & - & - & -  \\
    DEVA~\cite{cheng2023tracking} & R-50 & \cmark & 38.3 & - & - & 41.5 & - & - & - & - & - & - & - \\
    Tube-Link~\cite{tubelink} & R-50 & \cmark & 39.2 & - & - & 39.5 & 0.51 & 0.68 & 0.67 & 0.69 & 43.4 & 89.2 & 85.4  \\
    \textbf{\textsc{GvSeg}}   & R-50 & \cmark  & \textbf{44.0} & \textbf{44.4} & \textbf{42.4} & \textbf{44.9} & \textbf{0.53} & \textbf{0.71} & \textbf{0.69} & \textbf{0.71} & \textbf{44.5} & \textbf{90.5} & \textbf{86.4} \\

    \hline
    \color{mygray2}CFFM~\cite{sun2022coarse} & \color{mygray2}MiT-B5 & \color{mygray2}\xmark & \color{mygray2}- & \color{mygray2}- & \color{mygray2}- & \color{mygray2}- & \color{mygray2}- & \color{mygray2}- & \color{mygray2}- & \color{mygray2}- & \color{mygray2}49.3 & \color{mygray2}90.8 & \color{mygray2}87.1  \\
    \color{mygray2}MRCFA~\cite{sun2022mining} & \color{mygray2}MiT-B2 & \color{mygray2}\xmark & \color{mygray2}- & \color{mygray2}- & \color{mygray2}- & \color{mygray2}- & \color{mygray2}- & \color{mygray2}- & \color{mygray2}- & \color{mygray2}- & \color{mygray2}49.9 & \color{mygray2}90.9 & \color{mygray2}87.4  \\
    \color{mygray2}DVIS~\cite{zhang2023dvis}
    & \color{mygray2}Swin-L & \color{mygray2}\xmark & \color{mygray2}57.6 & \color{mygray2}59.9 & \color{mygray2}55.5 & \color{mygray2}55.3 & \color{mygray2}- & \color{mygray2}- & \color{mygray2}- & \color{mygray2}- & \color{mygray2}- & \color{mygray2}- & \color{mygray2}-  \\
    Video K-Net~\cite{li2022video} & Swin-B & \cmark & - & - & - & - & - & - & - & - & 57.2 & 90.1 & 87.8  \\
    TarVIS$^\dagger$~\cite{athar2023tarvis}
    & Swin-L & \cmark & 48.0 & 58.2 & 39.0 & 52.9 & - & - & - & - & - & - & -     \\
         DEVA~\cite{cheng2023tracking}
    & Swin-L & \cmark & 52.2 & - & - & 52.2 & - & - & - & - & - & - & -   \\
    Tube-Link~\cite{tubelink} 
    & Swin-B & \cmark & 50.4 & - & - & 49.4 & 0.56 & 0.72 & 0.69 & 0.74 & 62.3 & 91.4 & 89.3   \\
    \textbf{\textsc{GvSeg}}    & Swin-B& \cmark & \textbf{55.3} & \textbf{57.2} & \textbf{52.3} & \textbf{52.4} & \textbf{0.58} & \textbf{0.74} & \textbf{0.73} & \textbf{0.74}   & \textbf{63.2} & \textbf{91.8} & \textbf{89.4} \\
    \textbf{\textsc{GvSeg}}    & Swin-L& \cmark & \textbf{57.9} & \textbf{59.7} & \textbf{56.1} & \textbf{55.6} & - & - & - & -  & \textbf{65.5} & \textbf{93.8} & \textbf{91.6} \\
          \thickhline
      \end{tabular}  
   }
    
    \label{table:vps}
       \vspace{-15pt}
    \end{table*}

  \begin{table*}[t]
   \caption{
      Quantitative results for VIS on OVIS\!~\cite{qi2022occluded} and YouTube-VIS$_{21}$\!~\cite{yang2019video} (\S\ref{sec:VIS}). 
    }
    \vspace{-6pt}
    \resizebox{\linewidth}{!}{
          \setlength\tabcolsep{2.5pt}  
          \renewcommand\arraystretch{0.99}
          \begin{tabular}{r|c|c|ccccc|ccccccc}  
          \thickhline 
           \multirow{2}{*}{Method} & \multirow{2}{*}{Backbone} & {General} & \multicolumn{5}{c|}{Occluded-VIS \texttt{val}}  & \multicolumn{5}{c}{Youtube-VIS$_{21}$ \texttt{val}} \\
    \cline{4-14} 
          &                   & Solution & AP   & AP$_{\rm 50}$ & AP$_{\rm 75}$ & AR$_{\rm 1}$  & AR$_{\rm 10}$   &  AP  & AP$_{\rm 50}$ & AP$_{\rm 75}$ & AR$_{\rm 1}$  & AR$_{\rm 10}$ \\
    \hline
    \color{mygray2}SipMask~\cite{cao2020sipmask}     & \color{mygray2}R-50 & \color{mygray2}\xmark & \color{mygray2}\color{mygray2}10.2 & \color{mygray2}24.7 & \color{mygray2}7.8 & \color{mygray2}7.9 & \color{mygray2}15.8 &  
    \color{mygray2}31.7 & \color{mygray2}52.5 & \color{mygray2}34.0 & \color{mygray2}30.8 & \color{mygray2}37.8   \\
    \color{mygray2}InsPro~\cite{he2022inspro} & \color{mygray2}R-50 & \color{mygray2}\xmark & \color{mygray2}-   &  \color{mygray2}-   &  \color{mygray2}-   &  \color{mygray2}-   &  \color{mygray2}- & 
    \color{mygray2}37.6 & \color{mygray2}58.7 & \color{mygray2}0.9 & \color{mygray2}32.7 & \color{mygray2}41.4 \\
    \color{mygray2}SeqFormer~\cite{wu2022seqformer} & \color{mygray2}R-50 & \color{mygray2}\xmark & \color{mygray2}-   &  \color{mygray2}-   &  \color{mygray2}-   &  \color{mygray2}-   &  \color{mygray2}- & 
    \color{mygray2}40.5 & \color{mygray2}62.4 & \color{mygray2}43.7 & \color{mygray2}36.1 & \color{mygray2}48.1 \\
    \color{mygray2}VITA~\cite{heo2022vita} & \color{mygray2}R-50 & \color{mygray2}\xmark & \color{mygray2}19.6 & \color{mygray2}41.2 & \color{mygray2}17.4 & \color{mygray2}11.7 & \color{mygray2}26.0 &  
    \color{mygray2}45.7 & \color{mygray2}67.4 & \color{mygray2}49.5 & \color{mygray2}40.9 & \color{mygray2}53.6   \\
    \color{mygray2}MinVIS~\cite{huang2022minvis} & \color{mygray2}R-50 & \color{mygray2}\xmark & \color{mygray2}25.0 & \color{mygray2}45.5 & \color{mygray2}24.0 & \color{mygray2}13.9 & \color{mygray2}29.7 &  
    \color{mygray2}44.2 & \color{mygray2}66.0 & \color{mygray2}48.1 & \color{mygray2}39.2 & \color{mygray2}51.7   \\
    \color{mygray2}IDOL~\cite{wu2022defense} & \color{mygray2}R-50 & \color{mygray2}\xmark & \color{mygray2}30.2 & \color{mygray2}51.3 & \color{mygray2}30.0 & \color{mygray2}15.0 & \color{mygray2}37.5 &  
    \color{mygray2}43.9 & \color{mygray2}68.0 & \color{mygray2}49.6 & \color{mygray2}38.0 & \color{mygray2}50.9   \\
    \color{mygray2}MDQE~\cite{li2023mdqe} & \color{mygray2}R-50 & \color{mygray2}\xmark & \color{mygray2}33.0 & \color{mygray2}57.4 & \color{mygray2}32.2 & \color{mygray2}15.4 & \color{mygray2}38.4 &  
    \color{mygray2}44.5 & \color{mygray2}67.1 & \color{mygray2}48.7 & \color{mygray2}37.9 & \color{mygray2}49.8   \\
    \color{mygray2}DVIS~\cite{zhang2023dvis} & \color{mygray2}R-50 & \color{mygray2}\xmark & \color{mygray2}34.1 & \color{mygray2}59.8 & \color{mygray2}32.3 &  \color{mygray2}15.9 & \color{mygray2}41.1 &  
    \color{mygray2}-   &  \color{mygray2}-   &  \color{mygray2}-   &  \color{mygray2}-   &  \color{mygray2}-   \\
    \color{mygray2}GenVIS~\cite{heo2023generalized}  & \color{mygray2}R-50 & \color{mygray2}\xmark & \color{mygray2}34.5 & \color{mygray2}59.4 & \color{mygray2}35.0 & \color{mygray2}16.6 & \color{mygray2}38.3 &  
    \color{mygray2}47.1 & \color{mygray2}67.5 & \color{mygray2}51.5 & \color{mygray2}41.6 & \color{mygray2}54.7   \\
    \color{mygray2}TCOVIS~\cite{li2023tcovis} & \color{mygray2}R-50 & \color{mygray2}\xmark & \color{mygray2}35.3 & \color{mygray2}60.7 & \color{mygray2}36.6 & \color{mygray2}15.7 & \color{mygray2}39.5 &  
    \color{mygray2}49.5 & \color{mygray2}71.2 & \color{mygray2}53.8 & \color{mygray2}41.3 & \color{mygray2}55.9   \\
    \color{mygray2}CTVIS~\cite{ying2023ctvis} & \color{mygray2}R-50 & \color{mygray2}\xmark & \color{mygray2}35.5 & \color{mygray2}60.8 & \color{mygray2}34.9 & \color{mygray2}16.1 & \color{mygray2}41.9 &  
    \color{mygray2}50.1 & \color{mygray2}73.7 & \color{mygray2}54.7 & \color{mygray2}41.8 & \color{mygray2}59.5   \\
    TubeFormer~\cite{kim2022tubeformer} & R-50 & \cmark & -   &  -   &  -   &  -   &  - &
     41.2 & 60.4 & 44.7 & 40.4 & 54.0  \\
    CAROQ~\cite{choudhuri2023context} & R-50 & \cmark & 25.8 & 47.9 & 25.4 & 14.2 & 33.9 &  
    43.3 & 64.9 & 47.1 & 39.3 & 52.7   \\    
TarVIS~\cite{athar2023tarvis} & R-50 & \cmark & 31.1 & 52.5 & 30.4 & 15.9 & 39.9 &  
    48.3 & 69.6 & 53.2 & 40.5 & 55.9   \\
Tube-Link~\cite{tubelink} & R-50 & \cmark & 29.5 & 51.5 & 30.2 & 15.5 & 34.5 &  
    47.9 & 70.0 & 50.2 & 42.3 & 55.2   \\
    \textbf{\textsc{GvSeg}}                     & R-50 & \cmark  & \textbf{35.9} & \textbf{50.7} & \textbf{38.0} & \textbf{16.6} & \textbf{40.1}  & \textbf{49.6} & \textbf{72.0} & \textbf{53.1} & \textbf{42.7} & \textbf{56.7} \\
    \hline
    \color{mygray2}GenVIS~\cite{heo2023generalized} & \color{mygray2}Swin-L& \color{mygray2}\xmark & \color{mygray2}45.4 & \color{mygray2}69.2 & \color{mygray2}47.8 & \color{mygray2}18.9 & \color{mygray2}49.0 &  
    \color{mygray2}59.6 & \color{mygray2}80.9 & \color{mygray2}65.8 & \color{mygray2}48.7 & \color{mygray2}65.0   \\
    \color{mygray2}TCOVIS~\cite{li2023tcovis} & \color{mygray2}Swin-L& \color{mygray2}\xmark & \color{mygray2}46.7 & \color{mygray2}70.9 & \color{mygray2}49.5 & \color{mygray2}19.1 & \color{mygray2}50.8 &  
    \color{mygray2}61.3 & \color{mygray2}82.9 & \color{mygray2}68.0 & \color{mygray2}48.6 & \color{mygray2}65.1  \\
    \color{mygray2}CTVIS~\cite{ying2023ctvis} & \color{mygray2}Swin-L& \color{mygray2}\xmark & \color{mygray2}46.9 & \color{mygray2}71.5 & \color{mygray2}47.5 & \color{mygray2}19.1 & \color{mygray2}52.1 &  
    \color{mygray2}61.2 & \color{mygray2}84.0 & \color{mygray2}68.8 & \color{mygray2}48.0 & \color{mygray2}65.8   \\ 
    CAROQ~\cite{choudhuri2023context} & Swin-L& \cmark & -   &  -   &  -   &  -   &  - &
    54.5 & 75.4 & 60.5 & 45.5 & 61.4 \\
    TarVIS~\cite{athar2023tarvis} & Swin-L& \cmark & 43.2 & 67.8 & 44.6 & 18.0 & 50.4 &  
    60.2 & 81.4 & 67.6 & 47.6 & 64.8   \\
    
    Tube-Link~\cite{tubelink} & Swin-L& \cmark & -   &  -   &  -   &  -   &  - &
    58.4 & 79.4 & 64.3 & 47.5 & 63.6 \\
    \textbf{\textsc{GvSeg}}    & Swin-L& \cmark       & \textbf{49.7} & \textbf{74.9} & \textbf{52.0} & \textbf{18.9} & \textbf{54.5}  & \textbf{60.7} & \textbf{82.9} & \textbf{69.7} & \textbf{47.5} & \textbf{65.7} \\
          \thickhline
      \end{tabular}  
   }
    
    \label{table:vis}
       \vspace{-5pt}
    \end{table*}

\subsection{Results for Video Panoptic Segmentation}\label{sec:VPS}
\noindent\textbf{Dataset.$_{\!}$} 
VIPSeg\!~\cite{miao2022large} provides $2,806$/$323$ videos in \texttt{train}/\texttt{test} splits which covers 232 real-world scenarios and $58$/$66$ thing/stuff classes. KITTI-STEP\!~\cite{weber2021step} is an urban street-view dataset with 12/9 videos for \texttt{train}/\texttt{val}.
It includes 19 semantic classes, with two of them (\textit{pedestrians} and \textit{cars}) having tracking IDs.

\noindent\textbf{Evaluation$_{\!}$ Metric.$_{\!}$} Following conventions\!~\cite{miao2022large,weber2021step,kim2022tubeformer}, we adopt VPQ and STQ as metrics. VPQ computes the average panoptic quality from tube IoU across a span of several frames. For VIPSeg\!~\cite{miao2022large}, we further report the VPQ scores for \textit{thing} and \textit{stuff} classes (\ie, VPQ$^\text{Th}$ and VPQ$^\text{St}$). For KITTI-VPS\!~\cite{weber2021step}, we divide STQ into segmentation quality (SQ) and association quality (AQ) which evaluate the pixel-level tracking and segmentation performance in a video clip.

\noindent\textbf{Performance.} As illustrated by Table~\ref{table:vps}}, \textsc{GvSeg} achieves dominant results on VIPSeg\!~\cite{miao2022large}, presenting an improvement up to \textbf{4.8\%/5.4\%} in terms of VPQ/STQ over the SOTA\!~\cite{tubelink} with ResNet-50 as backbone.  This reinforces our belief that accommodating task-oriented property into general video segmentation is imperative. Such an assertion gets further support on KITTI-STEP\!~\cite{weber2021step} that \textsc{GvSeg} outperforms all existing solutions by significant margins in STQ and AQ, which focus more on the coherent association of identical objects.

\subsection{Results for Video Semantic Segmentation}\label{sec:VSS}
\noindent\textbf{Dataset.$_{\!}$$_{\!}$} VSPW~\cite{miao2021vspw}$_{\!}$ has$_{\!}$ $2,806/343$ in-the-wild$_{\!}$ videos$_{\!}$ with$_{\!}$ $198,224/24,502$ frames$_{\!}$ for$_{\!}$ \texttt{train}/\texttt{val}, and$_{\!}$ provides$_{\!}$ pixel-level annotations$_{\!}$ for$_{\!}$ 124$_{\!}$ semantic$_{\!}$ categories.

\noindent\textbf{Evaluation$_{\!}$ Metric.$_{\!}$} Following the standard evaluation protocol~\cite{miao2021vspw,tubelink}, we adopt the mean Intersection-over-Union (mIoU), and mean video consistency (mVC) which evaluates the category consistency among a video clip containing 8/16 frames (\ie, mVC$_\text{8}$ and mVC$_\text{16}$) as metrics.

\noindent\textbf{Performance.$_{\!}$} As shown in Table~\ref{table:vps}, based on ResNet-50, \textsc{GvSeg} outperforms all competitors and achieves \textbf{44.5\%} mIoU. In particular, the \textbf{90.5\%}/\textbf{86.4\%} scores in terms of mVC$_8$/mVC$_{16}$ are comparable to MRCFA\!~\cite{sun2022mining} which utilizes Swin-B as the backbone and yields much higher mIoU. This suggests that, benefited by task-oriented temporal contrast learning, \textsc{GvSeg} can produce more consistent prediction across frames. When integrated with Swin-B, \textsc{GvSeg} demonstrates \textbf{0.9\%} gains over Tube-Link\!~\cite{tubelink}, confirming the superiority of our approach.
\subsection{Results for Video Instance Segmentation}\label{sec:VIS}
\noindent\textbf{Dataset.$_{\!}$} 
Occluded VIS\!~\cite{qi2022occluded} is specifically designed to tackle the challenging scenario of object occlusions. It consists of $607/140$ long videos with up to 292 frames for \texttt{train}/\texttt{val} and spans 25 object categories with a high density of instances.
YouTube-VIS$_{21}$\!~\cite{yang2019video} comprises $2,985/421$ high resolution videos for \texttt{train}/\texttt{val}. It extensively covers 40 object classes with  $8,171$ unique instances.

\noindent\textbf{Evaluation$_{\!}$ Metric.$_{\!}$} Following the official setup\!~\cite{yang2019video,qi2022occluded}, we report the mean average precision (mAP) by averaging multiple IoU scores with thresholds from 0.5 to 0.95 at step 0.05, and the average recall (AR) given 1/10 segmented instances per video (\ie, AR$_\text{1}$, AR$_\text{10}$). AP$_\text{50}$ and AP$_\text{75}$ with IoU thresholds at 0.5 and 0.75 are also employed for further analysis. 

\noindent\textbf{Performance.} From Table~\ref{table:vis} we can observe that \textsc{GvSeg} provides a considerable performance gain over existing methods on Occluded-VIS\!~\cite{qi2022occluded}. Notably, it outperforms the prior specalized/general solution SOTA CTVIS\!~\cite{ying2023ctvis}/TarVIS\!~\cite{athar2023tarvis} by \textbf{0.4\%}/\textbf{4.8\%} in terms of mAP with ResNet-50 as the backbone.
When adopting Swin-L, \textsc{GvSeg} showcases far better performance, achieving up to \textbf{49.7\%} mAP which earns an impressive \textbf{2.8\%} improvement against CTVIS. Moreover, we report performance on YouTube-VIS$_{21}$\!~\cite{yang2019video}. As seen, \textsc{GvSeg} surpasses the main rival (\ie, TarVIS), by \textbf{1.3\%/0.5\%} with ResNet-50/Swin-L as backbone. 

   \begin{table*}[t]
    \caption{
       Quantitative results for EVS on YouTube-VOS$_{18}$\!~\cite{xu2018youtube}, and BURST\!~\cite{athar2023burst} (\S\ref{sec:EVS}). 
     }
     \vspace{-6pt}
     \resizebox{\linewidth}{!}{
           \setlength\tabcolsep{3.1pt}
           \renewcommand\arraystretch{0.99}
           \begin{tabular}{r|c|c|ccccc|ccc}
           \thickhline
            \multirow{2}{*}{Method} & \multirow{2}{*}{Backbone} & {General} & \multicolumn{5}{c|}{YouTube-VOS$_{18}$~\texttt{val}~(Mask-guide)} & \multicolumn{3}{c}{BURST~\texttt{val}~(Point-guide)} \\
     \cline{4-11} 
           &                   & Solution & \ \ $\mathcal{G}$  \ \  & \ \ $\mathcal{J}_{s}$ \ \  & \ \ $\mathcal{F}_{s}$ \ \  & \ \ $\mathcal{J}_{u}$ \ \  & \ \ $\mathcal{F}_{u}$ \ \ & \ \ H$_\text{all}$ \ \ & \ \ H$_\text{com}$ \ \ & \ \ H$_\text{unc}$ \ \  \\
     \hline
     \color{mygray2}Box Tracker~\cite{arne2020trackeval} & \color{mygray2}R-50 & \color{mygray2}\xmark & \color{mygray2}- & \color{mygray2}- & \color{mygray2}- & \color{mygray2}- & \color{mygray2}- & \color{mygray2}12.7 & \color{mygray2}31.7  & \color{mygray2}7.9  \\
     \color{mygray2}STCN~\cite{cheng2021rethinking} & \color{mygray2}R-50& \color{mygray2}\xmark & \color{mygray2}83.0 & \color{mygray2}81.9 & \color{mygray2}86.5 & \color{mygray2}77.9 & \color{mygray2}85.7 & \color{mygray2}24.4 & \color{mygray2}44.0 & \color{mygray2}19.5\\
     \color{mygray2}XMem~\cite{cheng2022xmem} & \color{mygray2}R-50& \color{mygray2}\xmark & \color{mygray2}85.7 & \color{mygray2}84.6 & \color{mygray2}89.3 & \color{mygray2}80.2 & \color{mygray2}88.7 & \color{mygray2}32.3 & \color{mygray2}47.5 & \color{mygray2}28.6\\
     UNINEXT~\cite{yan2023universal} & R-50& \cmark &77.0& 76.8& 81.0 &70.8 & 79.4 & - & - & - \\
    TarVIS~\cite{athar2023tarvis} & R-50 & \cmark  &79.2 & 79.7 & 84.2 & 72.9 & 79.9& 30.9 & 43.2 & 27.8  \\
     \textbf{\textsc{GvSeg}}     & R-50 & \cmark  & \textbf{81.5} & \textbf{80.9} & \textbf{86.0} & \textbf{75.4} & \textbf{83.7}  & \textbf{35.9} & \textbf{49.6} & \textbf{32.7} \\ 
     \hline
     UNINEXT~\cite{yan2023universal} & ConvNeXt-L& \cmark & 78.1 & 79.1 & 83.5 & 71.0 &  78.9 & - & - & - \\
     TarVIS~\cite{athar2023tarvis} & Swin-L& \cmark  & 82.1 & 82.3& 86.5& 76.1& 83.5& 37.5 & 51.7 & 34.0   \\
     \textbf{\textsc{GvSeg}}    & Swin-L& \cmark       & \textbf{84.3} & \textbf{82.7} & \textbf{87.9} & \textbf{78.5} & \textbf{87.1}  & \textbf{40.9} & \textbf{55.5} & \textbf{36.3} \\ 
     \thickhline  
       \end{tabular}  
    }
     
     \label{table:evs}
        \vspace{-15pt}
     \end{table*}

\subsection{Results for Exemplar-guided Video Segmentation}\label{sec:EVS}
\noindent\textbf{Dataset.$_{\!}$} YouTube-VOS$_{18}$\!~\cite{xu2018youtube} includes 3, 471/474 videos for \texttt{train}/\texttt{val}. The videos are sampled at 30 FPS and annotated per 5 frames with multiple objects. BURST\!~\cite{athar2023burst} contains 500/993/1, 421 videos for \texttt{train}/\texttt{val}/\texttt{test}. It provides mask/point/bounding box as exemplars  and averages over 1000 frames per video.

\noindent\textbf{Evaluation$_{\!}$ Metric.$_{\!}$} For YouTube-VOS$_{18}$, we report region similarity$_{\!}$~($\mathcal{J}$) and contour accuracy$_{\!}$~($\mathcal{F}$)
at$_{\!}$ \textit{seen}$_{\!}$ and$_{\!}$ \textit{unseen}$_{\!}$ classes. For BURST, we assess
higher order tracking accuracy\!~\cite{luiten2021hota} on common (H$_\text{com}$) and uncommon (H$_\text{unc}$) classes.

\noindent\textbf{Performance.} 
To make a fair comparison with existing work which usually tests on BURST without training, we train \textsc{GvSeg} on YouTube-VOS$_{18}$ and randomly adopt mask or point exemplars as the guidance. Then the performance is evaluated with mask exemplar on YouTube-VOS$_{18}$ and point exemplar on BURST. As shown in Table \ref{table:evs},
\textsc{GvSeg} yields satisfactory performance on YouTube-VOS$_{18}$, \ie, surpassing the general counterpart (\ie, TarVIS\!~\cite{athar2023tarvis}) by \textbf{2.3\%}/\textbf{2.2\%} in terms of $\mathcal{G}$ score with ResNet-50/Swin-L as the backbone. 
We also provide the point-guided segmentation results on BURST. As seen, \textsc{GvSeg} surpasses current solutions by a large margin across all metrics. For instance, When compared with task-specialized approaches (\eg, XMem\!~\cite{cheng2022xmem}), our approach still earns \textbf{3.6\%} improvement. Note existing work has to adopt an additional offline model for mask prediction with given points, while our method natively supports points as the exemplar, contributing to the superiority in both efficiency and effectiveness.

\subsection{Qualitative Results}\label{sec:qualitative}
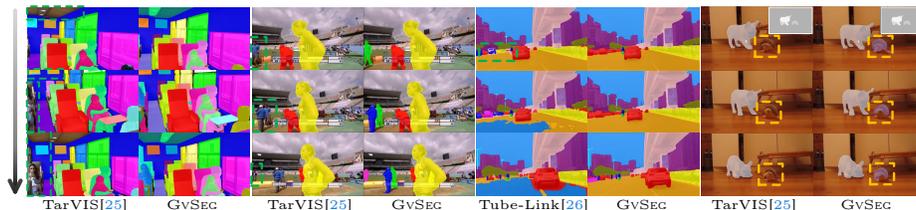
\begin{figure*}[t]
       \begin{center}
           \includegraphics[width=\linewidth]{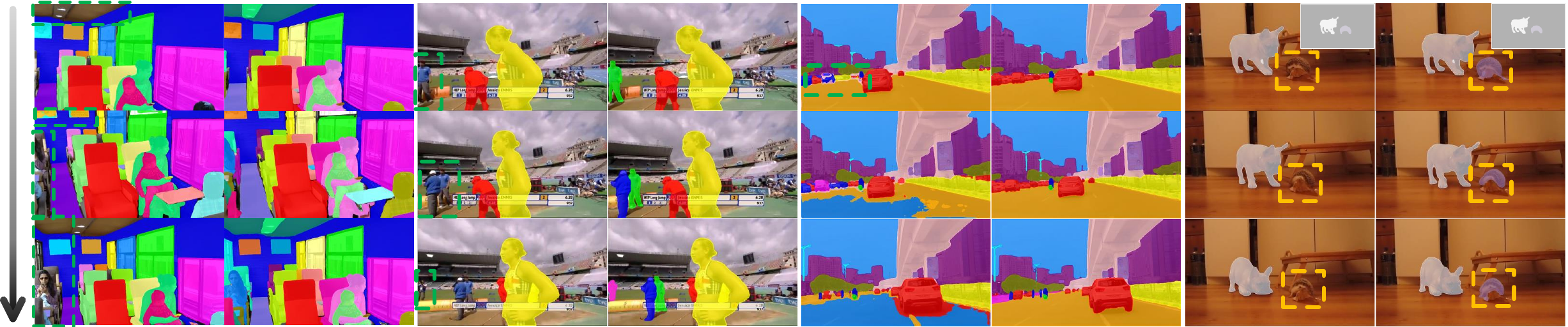}
           \put(-334,-4.6){{\tiny TarVIS\!\!~\cite{athar2023tarvis}}}
           \put(-287,-4.6){{\tiny \textsc{GvSeg}}}
           \put(-249,-4.6){{\tiny TarVIS\!\!~\cite{athar2023tarvis}}}
           \put(-202,-4.6){{\tiny \textsc{GvSeg}}}
           \put(-169,-4.6){{\tiny Tube-Link\!\!~\cite{tubelink}}}
           \put(-117,-4.6){{\tiny \textsc{GvSeg}}}
           \put(-81,-4.6){{\tiny TarVIS\!\!~\cite{athar2023tarvis}}}
           \put(-32,-4.6){{\tiny \textsc{GvSeg}}}
       \end{center}
       \vspace{-19pt}
       \captionsetup{font=small}
       \caption{\small \textbf{Visual comparison results} on VIPSeg-VPS\!~\cite{miao2022large}, YouTube-VIS$_{21}$\!~\cite{yang2019video}, VSPW-VSS\!~\cite{miao2021vspw} and YouTube-VOS$_{18}$\!~\cite{xu2018youtube} (\S\ref{sec:qualitative}).}
       \label{fig:results_youtube}
       \vspace{-5pt}
   \end{figure*}

In Fig.\!~\ref{fig:results_youtube}, we visualize the comparisons of \textsc{GvSeg} against the top-leading methods on four different tasks (\ie, VPS, VIS, VSS, and EVS). As seen, \textsc{GvSeg} gives more precise and consistent predictions in challenging scenarios.

\subsection{Diagnostic Experiment}\label{sec:diagnostic}
For more detailed analysis, we conduct a set of ablative studies on VIPSeg-VPS~\cite{miao2022large} with ResNet-50 as the backbone. 

\noindent\textbf{Key Component Analysis.} We investigate the improvements brought by each component of \textsc{GvSeg} in Table~\ref{table:ablation1} where `SPA' indicates `shape-position aware'. First, it can be observed that SPA query matching brings a considerable improvement over the Baseline, \ie, \textbf{1.8\%}/\textbf{1.2\%} concerning VPQ and STQ. This verifies our modeling of segment targets by disentangling them into appearance, shape, and position. Moreover, the adoption of task-oriented strategies for query initialization, object association, and temporal contrastive learning (TCL) elevates the results to a new level. Finally, we combine all these designs together which results in \textsc{GvSeg} and obtains the optimal performance. This confirms the compatibility of each component and the effectiveness of our whole algorithm.

\noindent\textbf{Matching Threshold \& Queue Length.} The results with different threshold $\tau$ and queue length $N_\mathcal{Q}$ utilized in task-oriented TCL are reported in Table~\ref{table:ablation5}. Though larger size of samples in the queue contributes to higher scores, we remain $N_\mathcal{Q}$ to 100 which gives nearly no impact in training speed and memory usage.

\begin{table*}[t]
  \caption{A set of ablative studies on VIPSeg-VPS~\cite{miao2022large} \texttt{val} with ResNet-50~\cite{he2016deep} as the backbone (\S\ref{sec:diagnostic}). The adopted settings are marked in red.}
     \vspace{-7pt}
  \hspace{-10pt}
  \begin{minipage}{.36\linewidth}
  \centering
  \subcaptionbox{$_{\!}$Component analysis \label{table:ablation1}}{
    \vspace{+2.5pt}
     \resizebox{\linewidth}{!}{
     \setlength\tabcolsep{1.4pt}
     \begin{tabular}{l|cc}
        \thickhline
        Component & VPQ $\uparrow$ & STQ $\uparrow$ \\ 
        \hline
        Baseline & 36.0 & 37.3 \\
        + SPA query matching & 37.8 & 38.5 \\
        + Task-oriented init.\&asso. & 40.1 & 40.7 \\
        + Task-oriented TCL & 41.2 & 42.0 \\
        \hline
        \ccnote{\textbf{\textsc{GvSeg}}} &\ccnote{\textbf{44.0}} & \ccnote{\textbf{44.9}}  \\
        \thickhline
     \end{tabular}
     }
  }
  \end{minipage}%
  \hfill%
  \begin{minipage}{.33\linewidth}
  \vspace{+0.5pt}
  \centering
  \captionsetup{font=small}
  \subcaptionbox{Task-oriented$_{\!}$ {TCL} \label{table:ablation5}}{
    \vspace{+3pt}
     \resizebox{\linewidth}{!}{
     \setlength\tabcolsep{9.4pt}
     \begin{tabular}{cc|cc}
        \thickhline
        $\tau$ & $N_\mathcal{Q}$ & VPQ $\uparrow$ & STQ $\uparrow$ \\
        \hline
        0.1 & 100 & 43.3 & 43.9 \\
        \arrayrulecolor{gray}\hdashline\arrayrulecolor{black}  
        \ccnote{\textbf{0.2}} & \ccnote{\textbf{100}} & \ccnote{\textbf{44.0}} & \ccnote{\textbf{44.9}} \\
        \arrayrulecolor{gray}\hdashline\arrayrulecolor{black}  
        0.2 & 200 & 44.1 & 45.1 \\
        0.3 & 100 & 43.6 & 44.4 \\
        0.3 & 200 & 43.7 & 44.6 \\
        \thickhline
     \end{tabular}
     }
  }
  \end{minipage}%
  \hfill%
  \begin{minipage}{.32\linewidth}
  \centering
  \subcaptionbox{Shape-position$_{\!}$ descriptor \label{table:ablation4}}{
     \vspace{+2.8pt}
     \resizebox{\linewidth}{!}{
     \setlength\tabcolsep{4.0pt}
     \begin{tabular}{cc|cc}
        \thickhline
        Angle $u$ & Radius $v$ & VPQ $\uparrow$ & STQ $\uparrow$ \\
        \hline
        12 & 6 & 43.1 & 43.8 \\
        24 & 12 & 43.6 & 44.3 \\
        \arrayrulecolor{gray}\hdashline\arrayrulecolor{black}  
        \ccnote{\textbf{36}} & \ccnote{\textbf{12}} & \ccnote{\textbf{44.0}} & \ccnote{\textbf{44.9}} \\
        \arrayrulecolor{gray}\hdashline\arrayrulecolor{black}  
        36 & 18 & 43.9 & 45.0 \\
        48 & 12 & 44.0 & 44.8 \\
        \thickhline
     \end{tabular}
     }
  }
  \end{minipage}
  
  \vspace{-0.7em} 

    \hspace{-7pt}
  \begin{minipage}{.56\linewidth}
  \centering
  \subcaptionbox{Task-oriented query association \label{table:ablation2}}{
     \resizebox{\linewidth}{!}{
     \setlength\tabcolsep{2.85pt}
     \begin{tabular}{c|cc|cc|cc}
        \thickhline
        \multirow{2}{*}{\#} &\multicolumn{2}{c|}{\textit{Thing}} &  \multicolumn{2}{c|}{\textit{Stuff}} &  \multirow{2}{*}{VPQ  $\uparrow$ } &  \multirow{2}{*}{STQ  $\uparrow$ } \\ \cline{2-5}
        &Appear. &Shape \& Pos. &Appear. &Shape \& Pos.& &    \\
        \hline
        1&\cmark& &  \cmark & & 42.1 & 43.1  \\
        \arrayrulecolor{gray}\hdashline\arrayrulecolor{black}  
        2&{\color{red}\cmark}& {\color{red}\cmark}  &{\color{red}\cmark} & & \ccnote{\textbf{44.0}} & \ccnote{\textbf{44.9}} \\
        3& \cmark&  & \cmark &  \cmark  & 41.7 & 42.8 \\ 
        4&\cmark & \cmark & \cmark  & \cmark  & 42.9 & 43.4 \\
        \thickhline
     \end{tabular}
     }
  }
  \end{minipage}%
  \hfill%
  \begin{minipage}{.43\linewidth}
  \centering
  \subcaptionbox{Task-oriented example sampling \label{table:ablation3}}{
     \resizebox{\linewidth}{!}{
     \setlength\tabcolsep{2.95pt} 
     \begin{tabular}{c|cc|cc|cc}
      \thickhline
      \multirow{2}{*}{\#} &\multicolumn{2}{c|}{\textit{Thing}} &  \multicolumn{2}{c|}{\textit{Stuff}} &  \multirow{2}{*}{VPQ  $\uparrow$ } &  \multirow{2}{*}{STQ  $\uparrow$ } \\ \cline{2-5}
      &Frame &Video&Frame &Dataset& &    \\
      \hline
      1&\cmark& & \cmark &  & 40.1 & 40.7  \\
      \arrayrulecolor{gray}\hdashline\arrayrulecolor{black}  
      2& \cmark& & &  \cmark  & 42.4 & 43.3 \\ 
      3&& \cmark & \cmark &  & 43.0 & 43.9 \\
      \arrayrulecolor{gray}\hdashline\arrayrulecolor{black}       
      4&& {\color{red}\cmark} &   &{\color{red}\cmark}  & \ccnote{\textbf{44.0}} & \ccnote{\textbf{44.9}} \\
      \thickhline
     \end{tabular}
     }
  }
  \end{minipage}
  \vspace{-25pt}
  \end{table*}

\noindent\textbf{Histogram Size.} In Table~\ref{table:ablation4}, we investigate the impact of the number of bins within the polar-style histogram for building position-shape descriptor. As seen, there is minor change in performance if $u\!\times\!v$ is large enough (\eg, $>200$) to capture the fine-grained variation in shape and location. 

\noindent\textbf{Task-Oriented Object Association.} We probe the impact of integrating distinct cues into object association in Table~\ref{table:ablation2}. By comparing \textit{Row} \#2 to \#1 we can observe that considering shape and position can boost the performance for \textit{thing} objects. In stark contrast, the inclusion of these cues causes negative impacts and yields less favorable results for \textit{stuff} objects (\ie, \textit{Row} \#3 \vs \#1).
This proves the necessity and urgency to cater to the task-oriented property which emphasizes more on \textit{instance discrimination} or \textit{semantic understanding}. 

\noindent\textbf{Task-Oriented Example Sampling.} To determine the contribution of our devised example sampling strategy utilized in TCL, we examine the performance  \wrt \textit{thing} and \textit{stuff} categories in Table~\ref{table:ablation3} where `Frame' refers to selecting samples from nearby frames, `Video' indicates gathering samples across the entire video based on shape-position descriptor for instance discrimination, and `Dataset' means storing samples in a queue to enhance the comprehension of semantics. As seen, both `Video' and `Dataset' level sampling for \textit{thing} and \textit{stuff} classes boost the scores significantly. This verifies our core insight that current sampling strategy in TCL is sub-optimal, and we can improve it by rendering a more holistic modeling on segment targets to select richer and more suitable samples.
 
\section{Conclusion}
We present \textsc{GvSeg}, the first {general} video segmentation solution that accommodates task-oriented properties into model learning. To achieve this, we first render a holistic investigation on segment targets by disentangling them into three essential constitutes: {appearance}, {shape}, and {position}. Then, by adjusting the involvement of these three key elements in query initialization and object association, we realize customizable prioritization of \textit{instance discrimination} or \textit{semantic understanding} to address different tasks. Moreover, task-oriented temporal contrastive learning is proposed to accumulate a diverse range of informative samples that considers both local consistency and semantic understanding properties for tracking instances and semantic/background classes, respectively. In this manner, \textsc{GvSeg} offers tailored consideration for each individual task and consistently obtains top-leading results in four video segmentation tasks. 

\bibliographystyle{splncs04}
\bibliography{egbib}

\newpage
\onecolumn
  \null
  \vskip .375in
  \begin{center}
    {\Large \bfseries Supplemental Material \par}
\end{center}

\appendix
\renewcommand{\thesection}{\Alph{section}}
\renewcommand{\thefigure}{S\arabic{figure}}
\renewcommand{\thetable}{S\arabic{table}}

\setcounter{section}{0}
\setcounter{figure}{0}
\setcounter{table}{0}

The appendix is \textbf{structured} as follows:
\begin{itemize}
   \vspace{-5pt}
   \setlength{\itemsep}{0pt}
   \setlength{\parsep}{0pt}
   \setlength{\parskip}{0pt}
   \item \S\ref{sec:ss1} provides more implementation details of \textsc{GvSeg}.
   \item \S\ref{sec:ss2} shows additional quantitative results on YouTube-VIS$_{19}$\!~\cite{yang2019video}. 
   \item \S\ref{sec:ss3} boardly discusses the Limitation, Boarder Impact and Future Work.
   \item \S\ref{sec:ss4} supplements more visualization results.
\end{itemize}

\vspace{-5pt}
\section{{More Implementation Details}}~\label{sec:ss1}
{GvSeg} is implemented on top of detectorn2. During training, for YouTube-VIS\!~\cite{yang2019video}/VOS\!~\cite{xu2018youtube}, the input frames are randomly cropped to ensure that the longer side is at most 768p/1024p for ResNet/Swin backbones, respectively. The shorter side is resized to at least 240p/360p and at most 480p/600p for ResNet/Swin. For OVIS\!~\cite{qi2022occluded}/VSPW\!~\cite{miao2021vspw}/VIPSeg\!~\cite{miao2022large}/KITTI\!~\cite{weber2021step}/BURST\!~\cite{athar2023burst},$_{\!}$ we$_{\!}$ resize$_{\!}$ the$_{\!}$ input$_{\!}$ frame$_{\!}$ so$_{\!}$ that$_{\!}$ the$_{\!}$ shorter$_{\!}$ side$_{\!}$ is$_{\!}$ at$_{\!}$ least$_{\!}$ 480p$_{\!}$ and$_{\!}$ at$_{\!}$ most$_{\!}$ 800p$_{\!}$ and the longer side is at most 1333p.

\vspace{-5pt}
\section{Additional Quantitative Results for VIS}\label{sec:ss2}
We provide additional results on YouTube-VIS$_{19}$\!~\cite{yang2019video} in Table \ref{table:19}. YouTube-VIS$_{19}$ consists of $2,238/343$ videos for \texttt{train}/\texttt{val}. Following official setting~\cite{yang2019video,qi2022occluded}, we adopt mean average precision (mAP) and average recall (AR) as evaluation metrics. The training settings remain consistent with those used for YouTube-VIS$_{21}$. We observed that \textsc{GvSeg} consistently outperforms previous state-of-the-art methods in terms of mAP and AR.

\begin{table}
    \centering
    \small
            \caption{\small\textbf{Quantitative results}$_{\!}$~on$_{\!}$ YouTube-VIS$_{19}$~\cite{yang2019video}$_{\!}$ \texttt{val} (\S\ref{sec:ss2}).}
            \vspace{-10pt}
    \resizebox{0.99\columnwidth}{!}{
        \setlength\tabcolsep{3.8pt}
        \renewcommand\arraystretch{0.99}
        \begin{tabular}{c|c|c|ccccc}
            \toprule[1.0pt]
            {Method} & {Backbone} & {Gen. Sol} & mAP & AP$_{\rm 50}$ & AP$_{\rm 75}$ & AR$_{\rm 1}$ & AR$_{\rm 10}$  \\ 
              \midrule
              \color{mygray2}MaskTrack~\cite{yang2019video}
            & \color{mygray2}R-50  & \color{mygray2}\xmark  & \color{mygray2}30.3 & \color{mygray2}51.1 & \color{mygray2}32.6& \color{mygray2}31.0 & \color{mygray2}35.5 \\
            \color{mygray2}SipMask~\cite{cao2020sipmask}
            & \color{mygray2}R-50 & \color{mygray2}\xmark & \color{mygray2}33.7 & \color{mygray2}54.1 & \color{mygray2}35.8 & \color{mygray2}35.4 & \color{mygray2}40.1  \\
            \color{mygray2}CrossVIS~\cite{yang2021crossover}
            & \color{mygray2}R-50 & \color{mygray2}\xmark & \color{mygray2}36.3 & \color{mygray2}56.8 & \color{mygray2}38.9 & \color{mygray2}35.6 & \color{mygray2}40.7  \\
            \color{mygray2}InsPro~\cite{he2022inspro}
            & \color{mygray2}R-50 & \color{mygray2}\xmark & \color{mygray2}37.6 & \color{mygray2}58.7 & \color{mygray2}0.9 & \color{mygray2}32.7 & \color{mygray2}41.4 \\
            \color{mygray2}VISOLO~\cite{han2022visolo} 
            & \color{mygray2}R-50 & \color{mygray2}\xmark & \color{mygray2}38.6 & \color{mygray2}56.3 & \color{mygray2}43.7 & \color{mygray2}35.7 & \color{mygray2}42.5 \\
            \color{mygray2}InstMove~\cite{liu2023instmove}
            & \color{mygray2}R-50 & \color{mygray2}\xmark & \color{mygray2}40.6 & \color{mygray2}67.2 & \color{mygray2}45.1 & \color{mygray2}35.0 & \color{mygray2}48.2 \\
         \color{mygray2}SeqFormer~\cite{wu2022seqformer}
            & \color{mygray2}R-50 & \color{mygray2}\xmark & \color{mygray2}47.4 & \color{mygray2}69.8 & \color{mygray2}51.8 & \color{mygray2}45.4 & \color{mygray2}54.8 \\
         \color{mygray2}MinVIS~\cite{huang2022minvis}
            & \color{mygray2}R-50 & \color{mygray2}\xmark & \color{mygray2}47.4 & \color{mygray2}69.0 & \color{mygray2}52.1 & \color{mygray2}45.7 & \color{mygray2}55.7 \\
            \color{mygray2}IDOL~\cite{wu2022defense}
            & \color{mygray2}R-50 & \color{mygray2}\xmark & \color{mygray2}49.5 & \color{mygray2}74.0 & \color{mygray2}52.9 & \color{mygray2}47.7 & \color{mygray2}58.7 \\
            \color{mygray2}VITA~\cite{heo2022vita}
            & \color{mygray2}R-50 & \color{mygray2}\xmark & \color{mygray2}49.8 & \color{mygray2}72.6 & \color{mygray2}54.5 & \color{mygray2}49.4 & \color{mygray2}61.0 \\
            \color{mygray2}GenVIS~\cite{heo2023generalized}
            & \color{mygray2}R-50 & \color{mygray2}\xmark & \color{mygray2}50.0 & \color{mygray2}71.5 & \color{mygray2}54.6 & \color{mygray2}49.5 & \color{mygray2}59.7 \\
            \color{mygray2}TCOVIS~\cite{li2023tcovis}
            & \color{mygray2}R-50 & \color{mygray2}\xmark & \color{mygray2}49.5 & \color{mygray2}71.2 & \color{mygray2}53.8 & \color{mygray2}41.3 & \color{mygray2}55.9 \\
            \color{mygray2}CTVIS~\cite{ying2023ctvis}
            & \color{mygray2}R-50 & \color{mygray2}\xmark & \color{mygray2}50.1 & \color{mygray2}73.7 & \color{mygray2}54.7 & \color{mygray2}41.8 & \color{mygray2}59.5 \\
            Mask2Former~\cite{cheng2021mask2former}
            & R-50 & \cmark & 46.4 & 68.0 & 50.0 & - & - \\
            CAROQ~\cite{choudhuri2023context}
            & R-50 & \cmark & 46.7 & 70.4 & 50.9 & 45.7 & 55.9 \\
            TubeFormer~\cite{kim2022tubeformer}
            & R-50 & \cmark & 47.5 & 68.7 & 52.1 & 50.2 & 59.0  \\
            Tube-Link~\cite{tubelink}
            & R-50 & \cmark & 52.8 & 75.4 & 56.5 & 49.3 & 59.9 \\
            \midrule
            \textbf{\textsc{GvSeg}}
            & R-50 & \cmark & \textbf{54.9} & \textbf{76.6} & \textbf{60.1} & \textbf{50.6} & \textbf{63.0}  \\
             \bottomrule[1.0pt]
        \end{tabular}
    }
    \captionsetup{font=small}
    \vspace{-2pt}
    \label{table:19}
        \vspace{-13pt}
\end{table}

\vspace{-5pt}
\section{Discussion}
\label{sec:ss3}
\begin{figure*}[t]
            \vspace{-0pt}
                \begin{center}
                    \includegraphics[width=\linewidth]{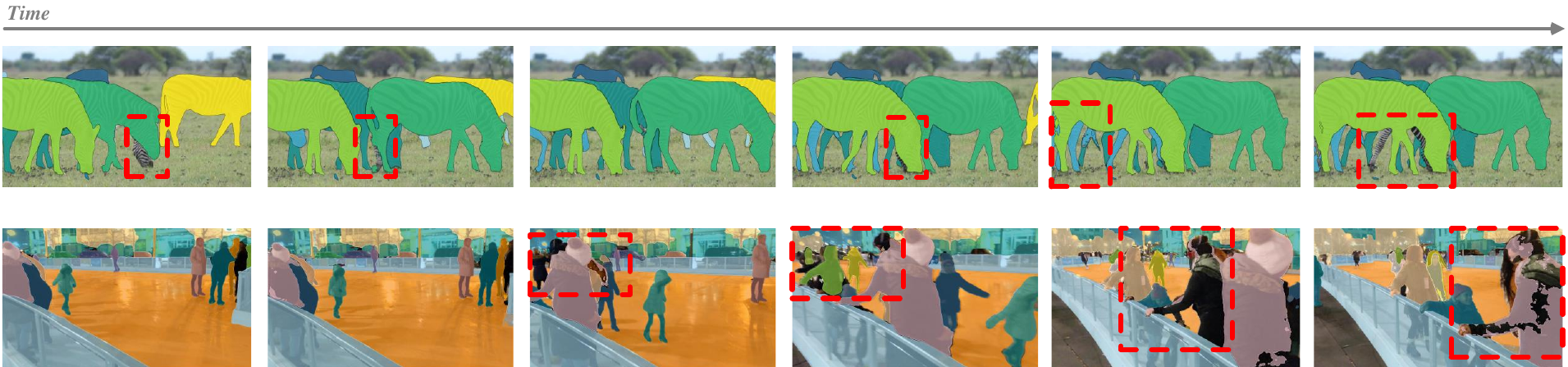}
                    \put(-344,34){{\scriptsize heavy occlusion (OVIS)}}
                    \put(-344,-6){{\scriptsize camera motion (VIPSeg)}}
                \end{center}
                \vspace{-10pt}
                \captionsetup{font=small}
                \caption{\small \textbf{Failure cases} due to on OVIS\!~\cite{qi2022occluded} and VIPSeg\!~\cite{miao2022large}. See more details in \S\ref{sec:ss3}.}
                \label{fig:failure}
                \vspace{-5pt}
            \end{figure*}

\noindent\textbf{Limitations.}
Although \textsc{GvSeg} has exhibited remarkable performance, environments with heavy occlusion and camera motion will result in subpar segmentation and tracking results. We show several representative failure cases in Fig.\!~\ref{fig:failure}. We aim to address these limitations in our future work.

\noindent\textbf{Broader Impact.}
Understanding visual scenes is a primary goal of computer vision. On the positive side, \textsc{GvSeg} represents a general video segmentation framework for EVS, VIS, VSS, and VPS which provides insight towards designing a universal model capable of addressing a broader spectrum of vision-related tasks. The disentanglement of task-specific properties of moving objects can benefit the wide application scenarios in video tasks such as Video Object Detection (VOD) and Multi-Object Tracking and Segmentation (MOTS). On the negative side, it's essential to acknowledge potential operational challenges our method may face in real-world applications. As a proactive step to mitigate any adverse effects on individuals and society, we advise the establishment of a robust security protocol which help ensure the safety and well-being of users and the broader community in case of any unforeseen issues.

\noindent\textbf{Future Work.}
Following the basic idea to disentangle task-specific properties of instances in a dynamic video, we will extend \textsc{GvSeg} towards a universal model with shared weights in our future work. We aim to cover more video instance perception tasks such as accommodate Single Object Tracking (SOT), Multi-Object Tracking and Segmentation (MOTS), Referring Expression Segmentation (RES), and Video Object Detection (VOD), all while maintaining shared weights across these tasks. This endeavor signifies a step towards a foundation model of video perception. In addition, while \textsc{GvSeg} emphasizes a unified architecture, the prospect of unified training is promising, and we shall consider it as our future direction.

\vspace{-5pt}
\section{Further Qualitative Results}~\label{sec:ss4}
\begin{figure*}[t]
   \vspace{-0pt}
       \begin{center}
           \includegraphics[width=\linewidth]{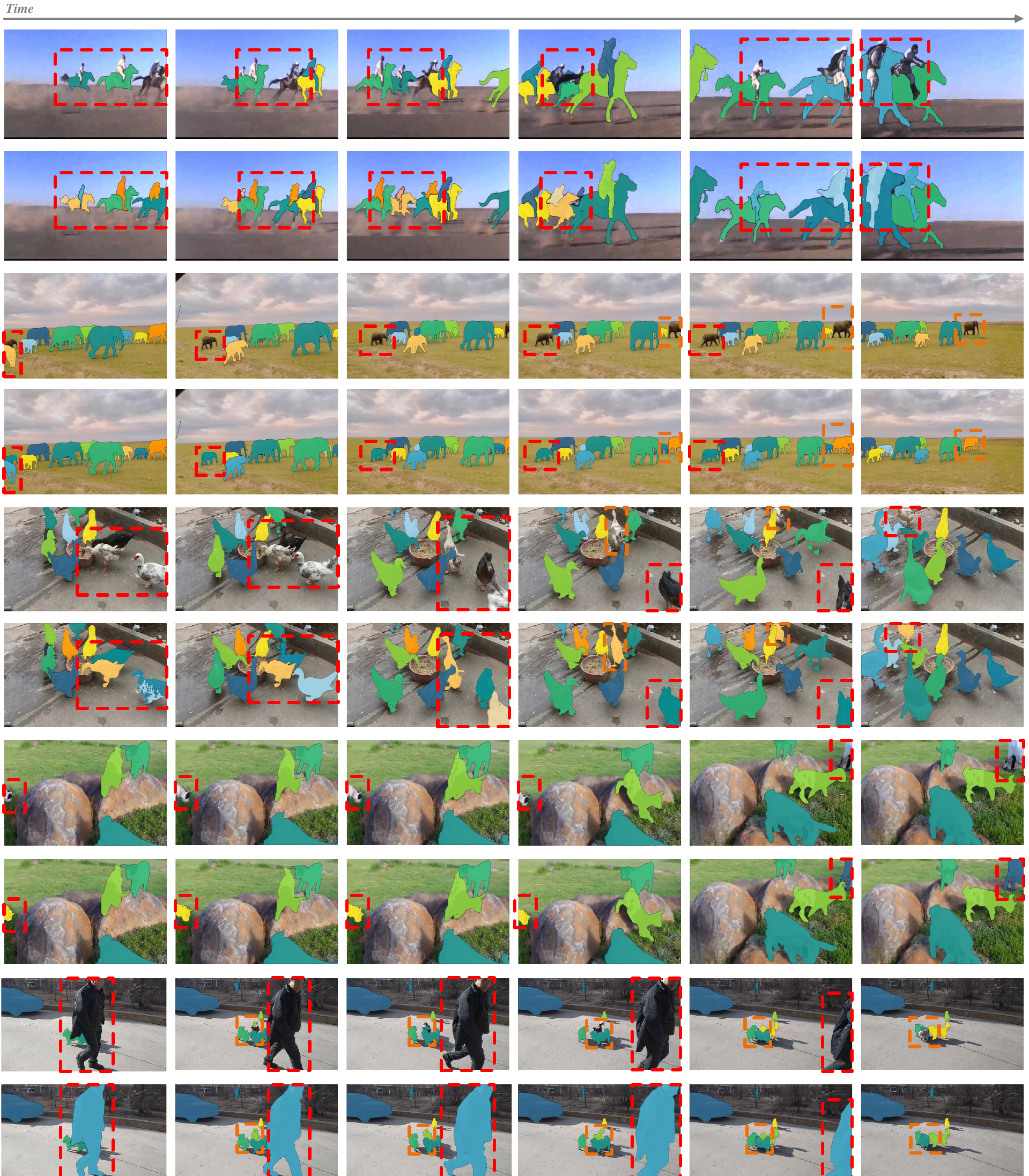}
           \put(-353,2){\rotatebox{90}{\scriptsize \textsc{GvSeg}}}
           \put(-353,38){\rotatebox{90}{\scriptsize TarVIS}}
           \put(-353,78){\rotatebox{90}{\scriptsize \textsc{GvSeg}}}
           \put(-353,116){\rotatebox{90}{\scriptsize TarVIS}}
           \put(-353,157){\rotatebox{90}{\scriptsize \textsc{GvSeg}}}
           \put(-353,196){\rotatebox{90}{\scriptsize TarVIS}}
           \put(-353,236){\rotatebox{90}{\scriptsize \textsc{GvSeg}}}
           \put(-353,274){\rotatebox{90}{\scriptsize TarVIS}}
           \put(-353,314){\rotatebox{90}{\scriptsize \textsc{GvSeg}}}
           \put(-353,357){\rotatebox{90}{\scriptsize TarVIS}}
       \end{center}
       \vspace{-10pt}
       \captionsetup{font=small}
       \caption{\small More \textbf{visual comparison} for Video Instance Segmentation on OVIS\!~\cite{qi2022occluded}.}
       \label{fig:results_ovis}
       \vspace{-5pt}
   \end{figure*}

\begin{figure*}[t]
      \vspace{-0pt}
          \begin{center}
              \includegraphics[width=\linewidth]{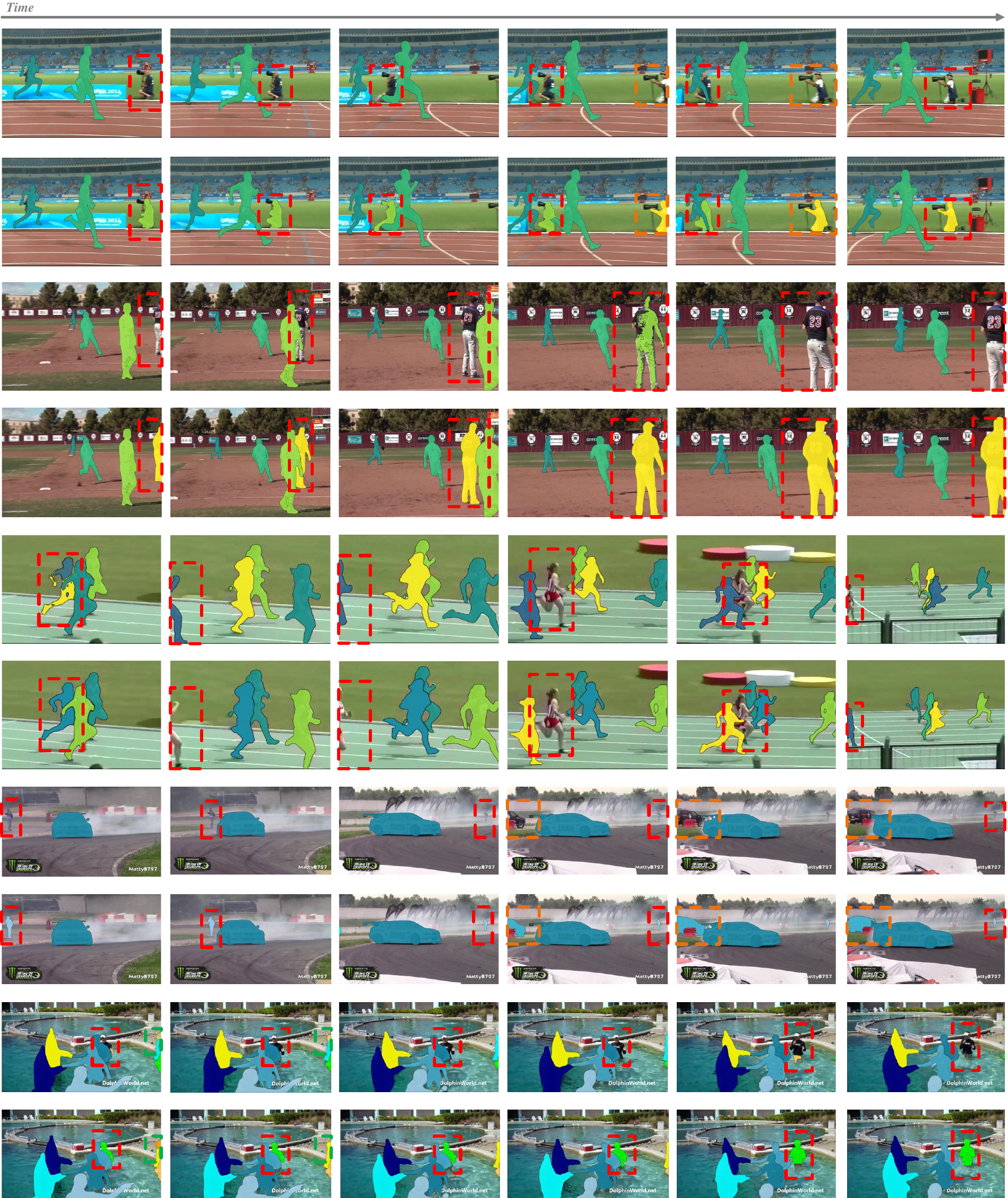}
              \put(-353,2){\rotatebox{90}{\scriptsize \textsc{GvSeg}}}
              \put(-353,39){\rotatebox{90}{\scriptsize TarVIS}}
              \put(-353,75){\rotatebox{90}{\scriptsize \textsc{GvSeg}}}
              \put(-353,113){\rotatebox{90}{\scriptsize TarVIS}}
              \put(-353,153){\rotatebox{90}{\scriptsize \textsc{GvSeg}}}
              \put(-353,196){\rotatebox{90}{\scriptsize TarVIS}}
              \put(-353,239){\rotatebox{90}{\scriptsize \textsc{GvSeg}}}
              \put(-353,284){\rotatebox{90}{\scriptsize TarVIS}}
              \put(-353,326){\rotatebox{90}{\scriptsize \textsc{GvSeg}}}
              \put(-353,369){\rotatebox{90}{\scriptsize TarVIS}}
          \end{center}
          \vspace{-10pt}
          \captionsetup{font=small}
          \caption{\small More \textbf{visual comparison} for Video Instance Segmentation on YouTube-VIS$_{21}$\!~\cite{yang2019video}.}
          \label{fig:results_youtube2}
          \vspace{-5pt}
      \end{figure*}

\begin{figure*}[t]
         \vspace{-0pt}
             \begin{center}
                 \includegraphics[width=\linewidth]{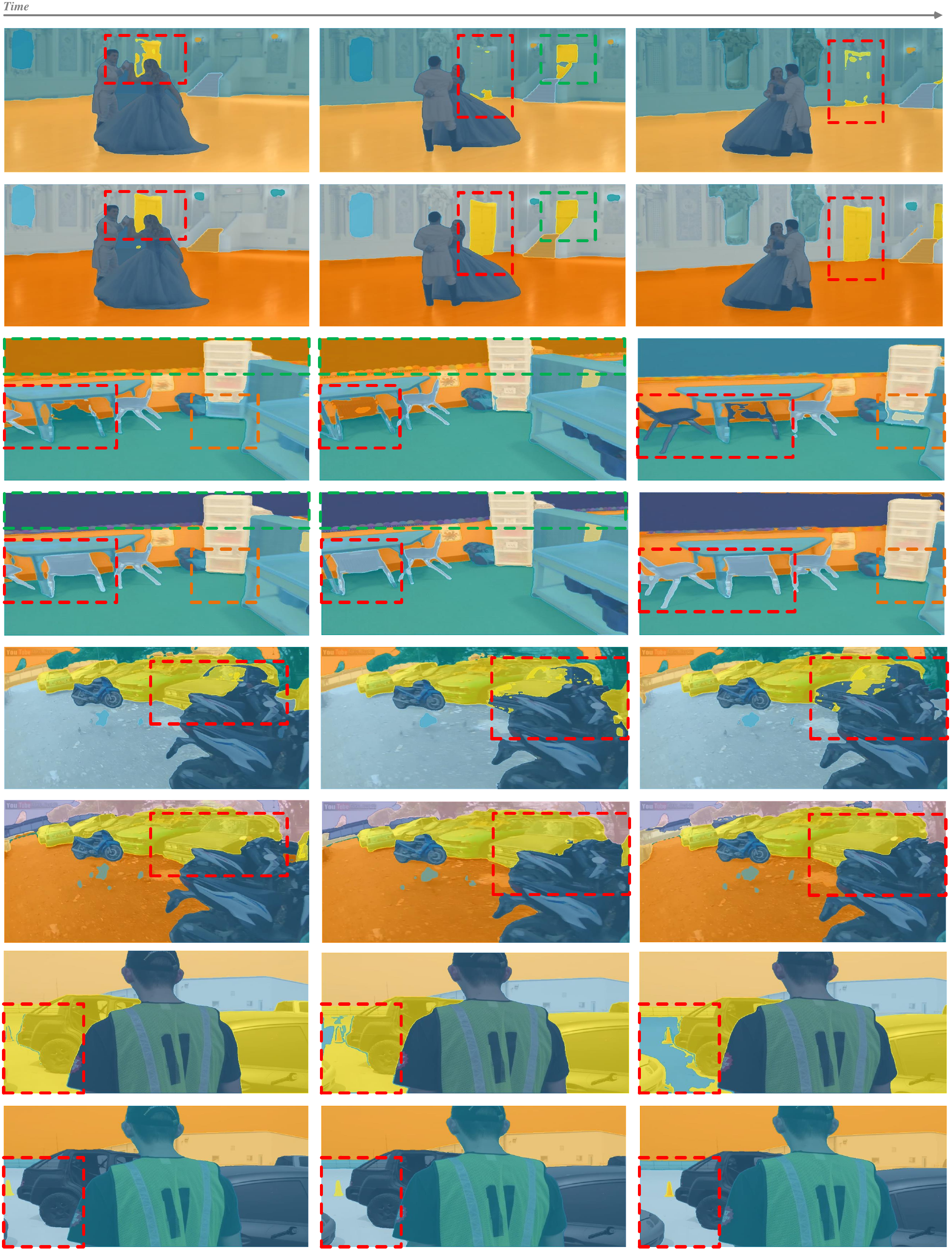}
                 \put(-353,13){\rotatebox{90}{\scriptsize \textsc{GvSeg}}}
                 \put(-353,65){\rotatebox{90}{\scriptsize Tube-Link}}
                 \put(-353,124){\rotatebox{90}{\scriptsize \textsc{GvSeg}}}
                 \put(-353,174){\rotatebox{90}{\scriptsize Tube-Link}}
                 \put(-353,234){\rotatebox{90}{\scriptsize \textsc{GvSeg}}}
                 \put(-353,286){\rotatebox{90}{\scriptsize Tube-Link}}
                 \put(-353,348){\rotatebox{90}{\scriptsize \textsc{GvSeg}}}
                 \put(-353,401){\rotatebox{90}{\scriptsize Tube-Link}}
             \end{center}
             \vspace{-10pt}
             \captionsetup{font=small}
             \caption{\small More \textbf{visual comparison} for Video Semantic Segmentation on VSPW\!~\cite{miao2021vspw}.}
             \label{fig:results_vspw}
             \vspace{-5pt}
         \end{figure*}

\begin{figure*}[t]
         \vspace{-0pt}
             \begin{center}
                 \includegraphics[width=\linewidth]{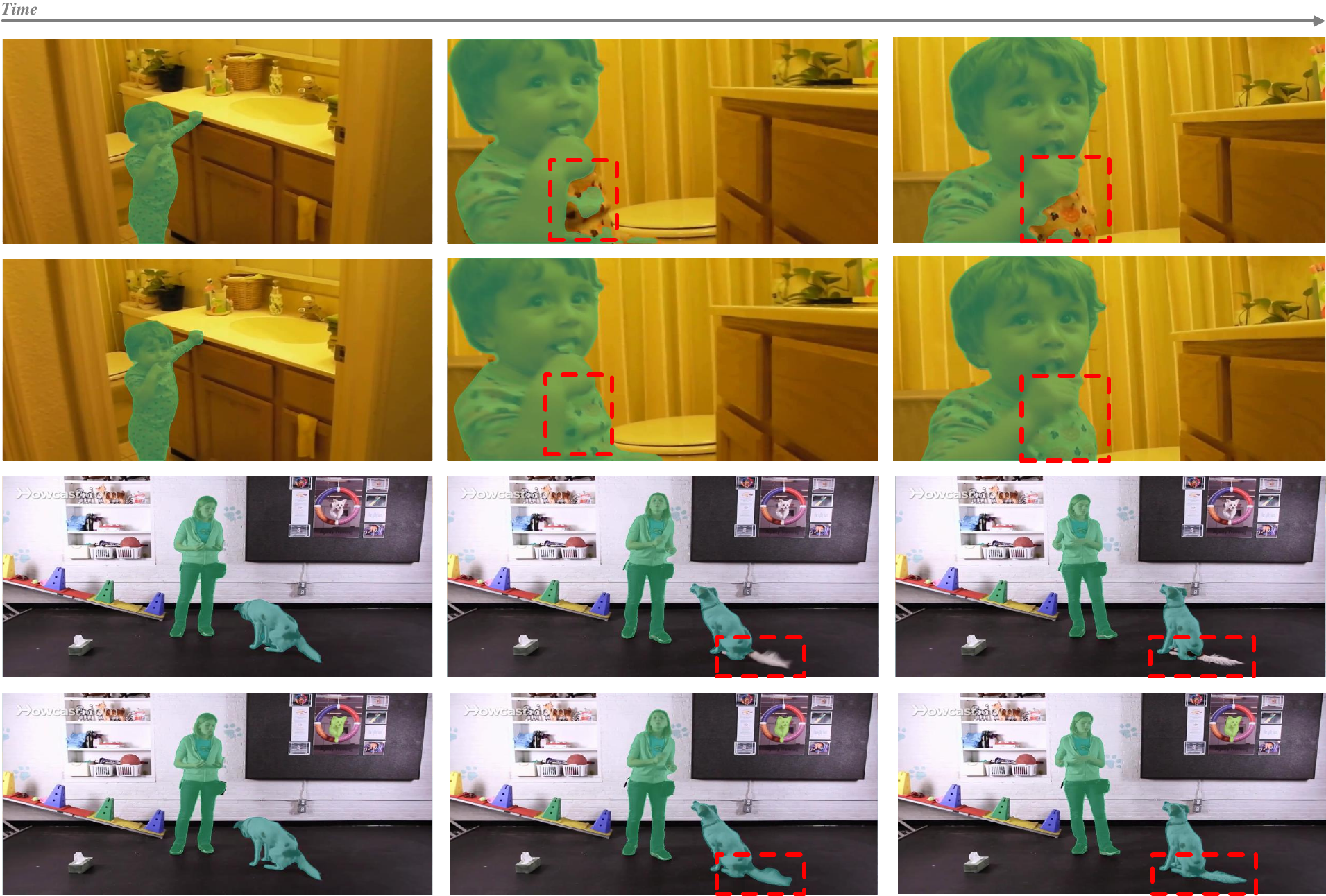}
                 \put(-353,14){\rotatebox{90}{\scriptsize \textsc{GvSeg}}}
                 \put(-353,70){\rotatebox{90}{\scriptsize TarVIS}}
                 \put(-353,126){\rotatebox{90}{\scriptsize \textsc{GvSeg}}}
                 \put(-353,182){\rotatebox{90}{\scriptsize TarVIS}}
             \end{center}
             \vspace{-10pt}
             \captionsetup{font=small}
             \caption{\small More \textbf{visual comparison} for Exemplar-guided Video Segmentation on BURST\!~\cite{athar2023burst}.}
             \label{fig:results_burst}
             \vspace{-5pt}
         \end{figure*}

\begin{figure*}[t]
         \vspace{-0pt}
             \begin{center}
                 \includegraphics[width=\linewidth]{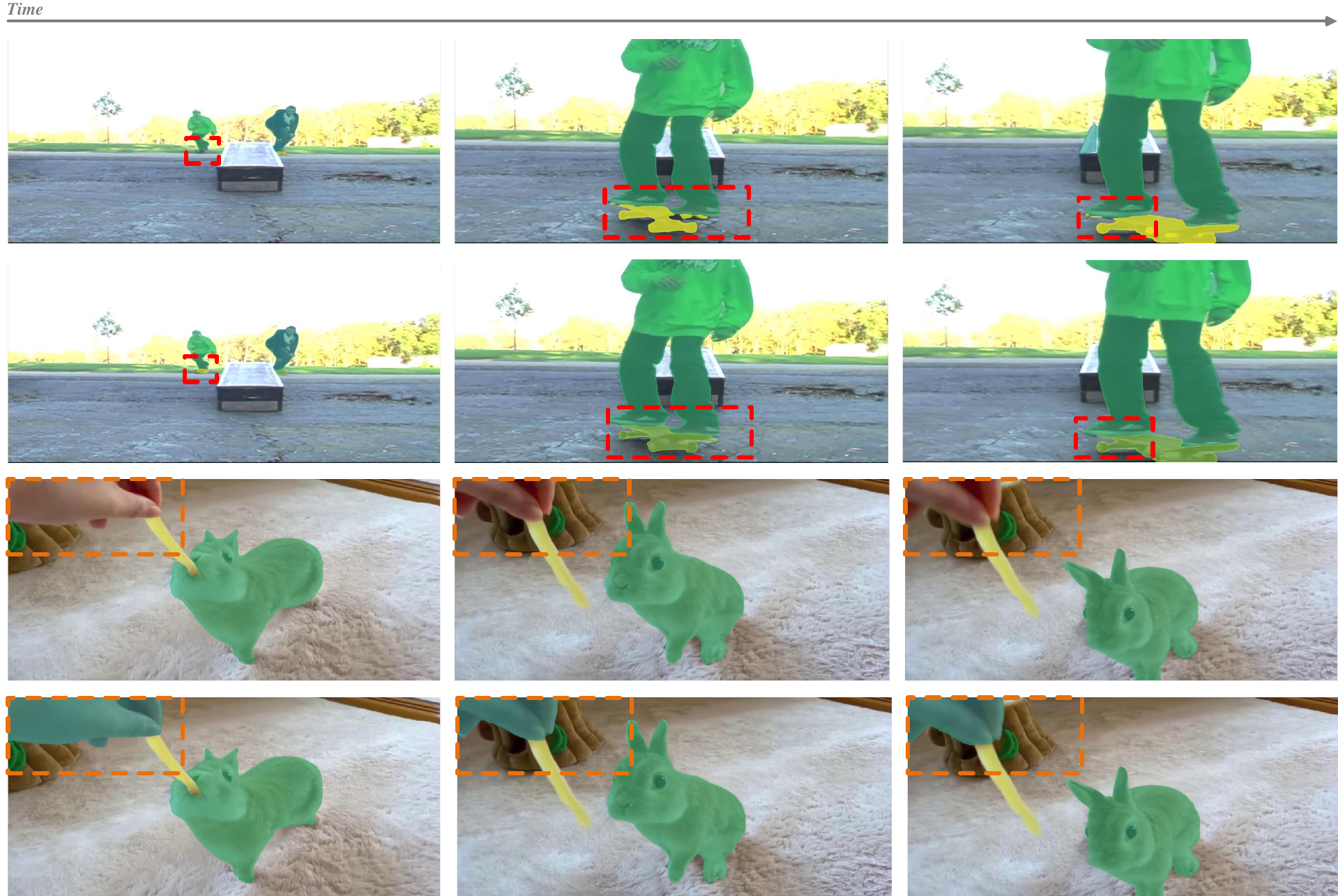}
                 \put(-353,14){\rotatebox{90}{\scriptsize \textsc{GvSeg}}}
                 \put(-353,68){\rotatebox{90}{\scriptsize TarVIS}}
                 \put(-353,126){\rotatebox{90}{\scriptsize \textsc{GvSeg}}}
                 \put(-353,180){\rotatebox{90}{\scriptsize TarVIS}}
             \end{center}
             \vspace{-10pt}
             \captionsetup{font=small}
             \caption{\small More \textbf{visual comparison} for Exemplar-guided Video Segmentation on YouTube-VOS\!~\cite{xu2018youtube}.}
             \label{fig:results_youtubevos2}
             \vspace{-5pt}
         \end{figure*}

\begin{figure*}[t]
      \vspace{-0pt}
          \begin{center}
              \includegraphics[width=\linewidth]{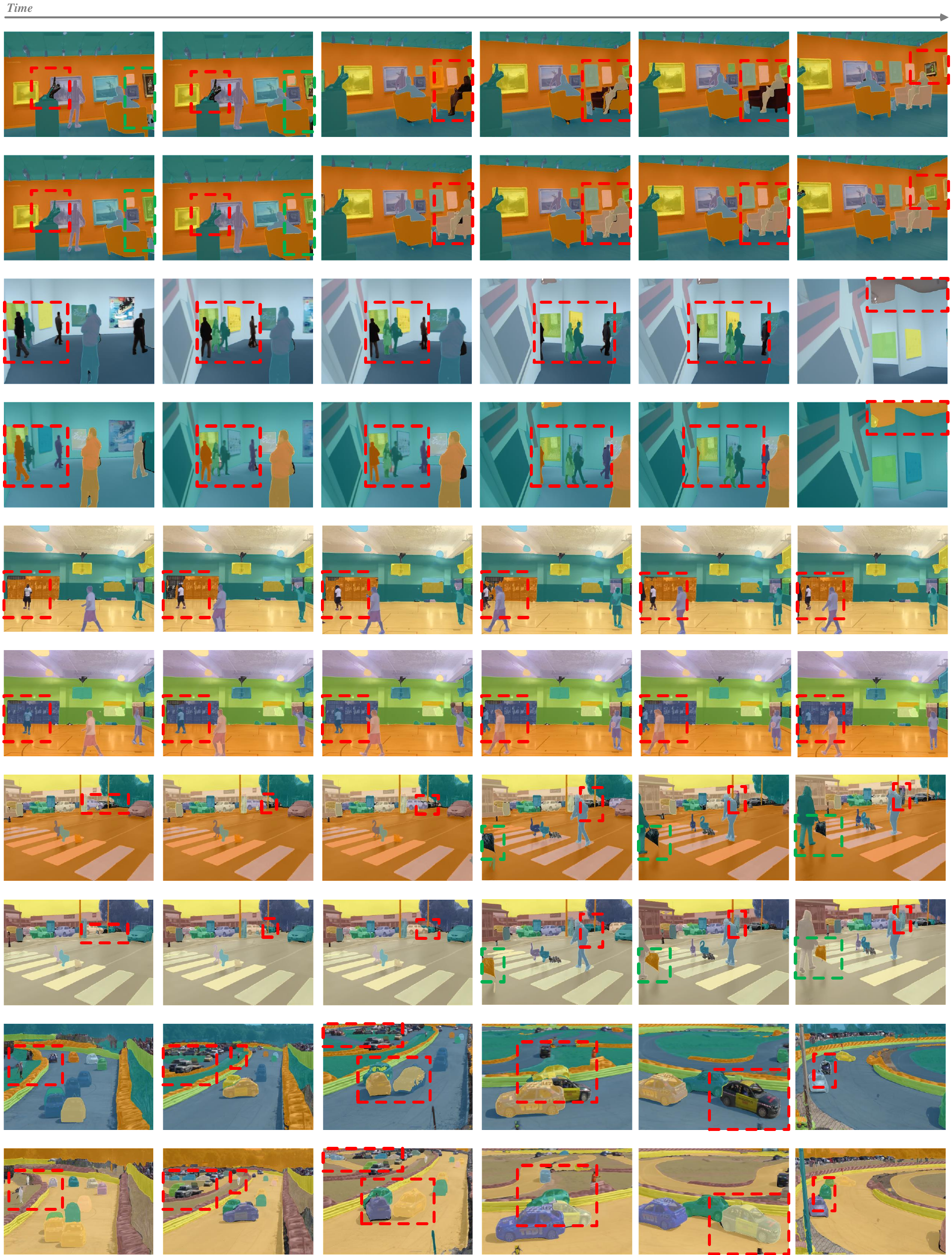}
              \put(-353,5){\rotatebox{90}{\scriptsize \textsc{GvSeg}}}
              \put(-353,52){\rotatebox{90}{\scriptsize TarVIS}}
              \put(-353,95){\rotatebox{90}{\scriptsize \textsc{GvSeg}}}
              \put(-353,143){\rotatebox{90}{\scriptsize TarVIS}}
              \put(-353,188){\rotatebox{90}{\scriptsize \textsc{GvSeg}}}
              \put(-353,234){\rotatebox{90}{\scriptsize TarVIS}}
              \put(-353,278){\rotatebox{90}{\scriptsize \textsc{GvSeg}}}
              \put(-353,324){\rotatebox{90}{\scriptsize TarVIS}}
              \put(-353,369){\rotatebox{90}{\scriptsize \textsc{GvSeg}}}
              \put(-353,415){\rotatebox{90}{\scriptsize TarVIS}}
          \end{center}
          \vspace{-10pt}
          \captionsetup{font=small}
          \caption{\small More \textbf{visual comparison} for Video Panoptic Segmentation on VIPSeg\!~\cite{miao2022large}.}
          \label{fig:vipseg}
          \vspace{-5pt}
      \end{figure*}

\begin{figure*}[t]
         \vspace{-0pt}
             \begin{center}
                 \includegraphics[width=\linewidth]{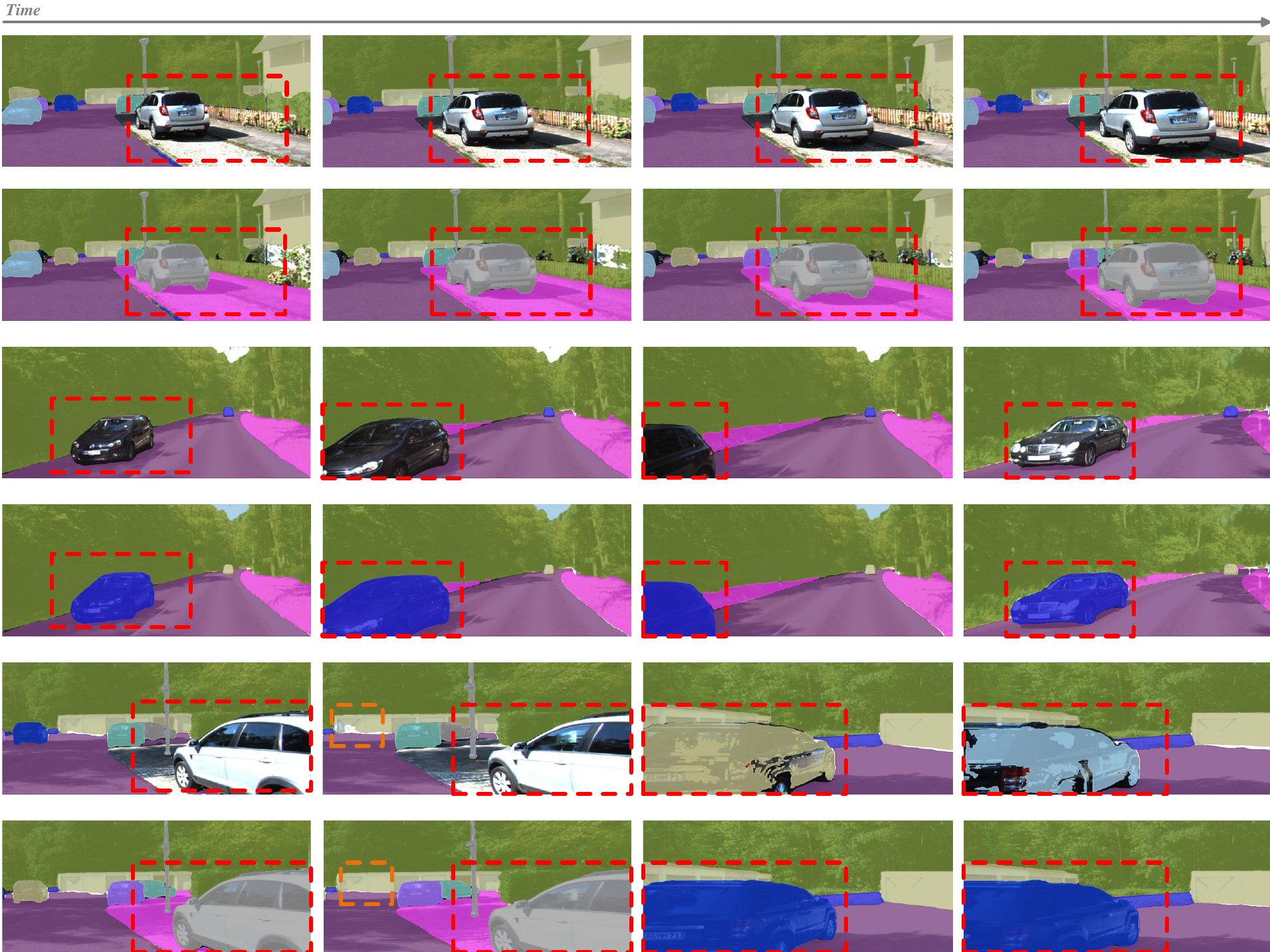}
                 \put(-353,5){\rotatebox{90}{\scriptsize \textsc{GvSeg}}}
                 \put(-353,48){\rotatebox{90}{\scriptsize TarVIS}}
                 \put(-353,91){\rotatebox{90}{\scriptsize \textsc{GvSeg}}}
                 \put(-353,135){\rotatebox{90}{\scriptsize TarVIS}}
                 \put(-353,179){\rotatebox{90}{\scriptsize \textsc{GvSeg}}}
                 \put(-353,220){\rotatebox{90}{\scriptsize TarVIS}}
             \end{center}
             \vspace{-10pt}
             \captionsetup{font=small}
             \caption{\small More \textbf{visual comparison} for Video Panoptic Segmentation on KITTI\!~\cite{weber2021step}.}
             \label{fig:results_kitti}
             \vspace{-5pt}
         \end{figure*}

In this section, we provide more qualitative results on five datasets, including OVIS\!~\cite{qi2022occluded} in Fig.\!~\ref{fig:results_ovis}, YouTube-VIS$_{21}$\!~\cite{yang2019video} in Fig.\!~\ref{fig:results_youtube2}, VSPW\!~\cite{miao2021vspw} in Fig.\!~\ref{fig:results_vspw}, BURST\!~\cite{athar2023burst} in Fig.\!~\ref{fig:results_burst}, YouTube-VOS\!~\cite{xu2018youtube} in Fig.\!~\ref{fig:results_youtubevos2} VIPSeg\!~\cite{miao2022large} in Fig.\!~\ref{fig:vipseg}, and KITTI\!~\cite{weber2021step} in Fig.\!~\ref{fig:results_kitti}. We observe that \textsc{GvSeg} is able to produce highly exquisite results compared with previous competitive methods TarVIS~\cite{athar2023tarvis} and Tube-Link~\cite{tubelink}.

\clearpage

\end{document}